\documentclass{article}

\usepackage{microtype}
\usepackage{graphicx}
\usepackage{subfigure}
\usepackage{booktabs} 
\usepackage{hyperref}

\usepackage[accepted]{icml2020}

\usepackage{hyperref}
\usepackage{url}
\usepackage{graphicx}
\usepackage{amsmath,amssymb,amsthm,amsfonts,mathrsfs,bm,multirow}
\usepackage{algorithm}
\usepackage{algorithmic}
\usepackage{psfrag}
\usepackage{grffile,xspace}
\usepackage{color, colortbl,bm}
\usepackage{accsupp}
\usepackage{adjustbox}
\usepackage{multirow}
\usepackage{booktabs}
\usepackage{verbatim}
\usepackage{xcolor}
\usepackage{tcolorbox}
\usepackage{wrapfig,lipsum}

\newcommand{\x}{\mathbf{x}}

\newcommand{\eps}{\epsilon}

\DeclareMathOperator*{\argmax}{arg\,max}

\newtheorem{prop}{Proposition}

\DeclareMathOperator*{\minimize}{\text{minimize}}
\DeclareMathOperator*{\maximize}{\text{maximize}}
\DeclareMathOperator*{\st}{\text{subject to}}
\DeclareMathAlphabet\mathbfcal{OMS}{cmsy}{b}{n}
\newcommand{\Def}[0]{\mathrel{\mathop:}=}

\newcommand{\mycomment}[1]{}
\newcommand{\first}[1]{#1}
\newcommand{\second}[1]{#1}
\newcommand{\third}[1]{#1}
\newcommand{\highlight}[1]{\textcolor{black}{\textbf{#1}}}

\icmltitlerunning{Proper Network Interpretability Helps Adversarial Robustness in Classification}
\begin{document}

\twocolumn[
\icmltitle{Proper Network Interpretability Helps Adversarial Robustness in Classification}




\begin{icmlauthorlist}
\icmlauthor{Akhilan Boopathy}{mit}
\icmlauthor{Sijia Liu}{ibm}
\icmlauthor{Gaoyuan Zhang}{ibm}
\icmlauthor{Cynthia Liu}{mit}
\icmlauthor{Pin-Yu Chen}{ibm}
\icmlauthor{Shiyu Chang}{ibm}
\icmlauthor{Luca Daniel}{mit}
\end{icmlauthorlist}

\icmlaffiliation{mit}{Massachusetts Institute of Technology}
\icmlaffiliation{ibm}{MIT-IBM Watson AI Lab, IBM Research}

\icmlcorrespondingauthor{Akhilan Boopathy, Sijia Liu}{akhilan@mit.edu, lsjxjtu@gmail.com}

\icmlkeywords{Machine Learning, ICML}

\vskip 0.3in
]



\printAffiliationsAndNotice{} 

\begin{abstract} 

Recent works have empirically shown that there exist   adversarial examples that can be hidden from neural network interpretability {(namely, making network interpretation maps visually similar)}, or  interpretability is itself susceptible to adversarial attacks. In this paper, we theoretically show  that with a proper measurement of interpretation, it is actually \textit{difficult} to prevent prediction-evasion adversarial attacks from causing interpretation discrepancy, 
as confirmed by experiments on MNIST, CIFAR-10 and Restricted ImageNet. Spurred by that,   we develop an interpretability-aware
defensive scheme built only on promoting robust interpretation (without the need for resorting to adversarial loss minimization). We show that our defense achieves both robust classification and robust interpretation,  outperforming state-of-the-art adversarial training methods   against   attacks of large perturbation in particular.   
\end{abstract}

\section{Introduction}
It has become widely known that convolutional neural networks (CNNs) are vulnerable to \textit{adversarial examples}, namely, perturbed inputs with the intention to mislead  networks' prediction \citep{szegedy2014intriguing,Goodfellow2015explaining,papernot2016limitations,carlini2017towards,chen2017ead,su2018robustness}. The  vulnerability of CNNs has spurred extensive research on adversarial  attack and defense. To design  adversarial attacks,  most works have focused on creating either imperceptible input perturbations \citep{Goodfellow2015explaining,papernot2016limitations,carlini2017towards,chen2017ead} or  adversarial patches robust to  the physical environment \citep{eykholt2018robust,brown2017adversarial,athalye2017synthesizing}.  Many defense methods have also  been developed to   prevent CNNs from misclassification when facing adversarial attacks. Examples include defensive  distillation \citep{papernot2016distillation}, training with adversarial examples \citep{Goodfellow2015explaining}, input gradient or curvature regularization \citep{ross2018improving,moosavi2019robustness}, adversarial training via robust optimization \citep{madry2017towards}, and TRADES to trade adversarial robustness off against accuracy \citep{zhang2019theoretically}. \textit{Different from the aforementioned works}, this paper attempts to understand the adversarial robustness of CNNs from the network interpretability perspective, and provides novel insights on when and how   interpretability could help robust classification. 

Having a prediction might not be enough {for many real-world machine learning applications}. 
It is  crucial to demystify why they 
make certain decisions. Thus, the problem of network interpretation arises. 
Various  methods have been proposed   to understand
the  mechanism of     decision making by CNNs.
One category of methods justify a prediction decision by assigning importance values   to reflect the influence of individual pixels or image sub-regions  on the final classification. Examples include pixel-space sensitivity map methods \citep{simonyan2013deep, zeiler2014visualizing, springenberg2014striving, smilkov2017smoothgrad,sundararajan2017axiomatic} and class-discriminative localization methods \citep{zhou2016learning,selvaraju2017grad,chattopadhay2018grad,petsiuk2018rise}, where the former evaluates the   sensitivity of a network classification decision to pixel variations at the input,  and the latter localizes which parts of an input image
were looked at by the network for making a classification decision. We refer readers to Sec.\,\ref{sec: preliminary} for some representative interpretation methods. Besides interpreting CNNs via feature importance maps, some  methods peek into the internal response of neural networks. Examples include   network dissection \citep{bau2017network}, and learning   perceptually-aligned    representations from adversarial training \citep{engstrom2019learning}.

Some recent works \citep{xu2018structured,xu2019interpreting,ZWJ+18,subramanya2018towards,ghorbani2019interpretation, dombrowski2019explanations,chen2019robust} began to study adversarial robustness by exploring the
spectrum between classification accuracy and network interpretability. It was shown in \citep{xu2018structured,xu2019interpreting} that an imperceptible adversarial perturbation  to fool classifiers can lead to  a significant change  in a class-specific network interpretation map.
Thus, it was argued that 
such an {interpretation discrepancy} can be used as a helpful metric to differentiate   adversarial examples from    benign   inputs. Nevertheless, the work \citep{ZWJ+18,subramanya2018towards} showed that under certain conditions, 
generating an attack  (which we call an \textit{interpretability sneaking attack, ISA}) 
that fools the classifier while keeping it stealthy from the coupled interpreter is 
\textit{not} significantly more difficult than generating an adversarial input that deceives the classifier only. Here   \textit{stealthiness} refers to keeping the interpretation map of an adversarial example highly similar to that of the corresponding benign example. The existing work had no {agreement} on the relationship between  robust classification and   network interpretability. In this work, we will revisit the validity of ISA and propose a     solution to improve the adversarial robustness of CNNs by leveraging robust interpretation in a proper way. 

The most relevant work to ours is \citep{chen2019robust}, which  
proposed a  robust attribution training  method with the aid of integrated gradient (IG), an axiomatic attribution map. It showed that the robust attribution training provides a   generalization of several commonly-used robust training methods  to defend adversarial attacks.  

Different from the previous work, our paper contains the following contributions. 
\begin{enumerate}
    \item By revisiting the validity of ISA, 
    we show that enforcing stealthiness of adversarial examples to a network interpreter could be challenging. Its difficulty
    relies on how one measures the interpretation discrepancy caused by input perturbations.
    \item We propose an $\ell_1$-{norm}  2-class interpretation discrepancy measure and  theoretically show that constraining it helps adversarial robustness. 
    {Spurred by that, we develop a principled interpretability-aware robust training method, which provides a means to achieve robust classification  by robust interpretation directly.}
    \item We empirically show that interpretability alone can be used to defend  adversarial attacks for both misclassifcation and    misinterpretation.
    Compared to the IG-based robust attribution training 
    \citep{chen2019robust},  our approach is lighter in computation 
    and provides 
   better robustness even when facing a strong adversary. 
\end{enumerate}

\section{Preliminaries and Motivation}\label{sec: preliminary}

In this section, we provide a brief background on interpretation methods of CNNs for justifying a classification decision, and motivate  
 the phenomenon  {of}  \textit{interpretation discrepancy}  caused by adversarial examples.

To explain what and why CNNs predict, 
we consider two types of network interpretation methods: a) \textit{class activation map (CAM)} \citep{zhou2016learning,selvaraju2017grad,chattopadhay2018grad} and b)    \textit{pixel sensitivity map (PSM)}  \citep{simonyan2013deep,springenberg2014striving,smilkov2017smoothgrad,sundararajan2017axiomatic,yeh2019infidelity}. 
Let  $f (\mathbf x) \in \mathbb R^C$ denote a CNN-based predictor   that maps an input $\mathbf x \in \mathbb R^d$ to a probability
vector of   $C$ classes. Here $f_c(\mathbf x)$, the $c$th element of $f(\mathbf x)$, denotes the classification score  (given by logit before the softmax) for class $c$.  Let $I(\mathbf x, c)  $ denote an interpreter   (CAM or PSM) that reflects where in $\mathbf x$ contributes to the classifier's decision on $c$.

\paragraph{CAM-type methods.}  CAM \citep{zhou2016learning} produces a  class-discriminative  localization map for   CNNs, which performs  global averaging pooling over convolutional feature maps prior to the softmax. Let the penultimate layer output $K$ feature maps, each of which is denoted by a vector representation ${\mathbf A}_k \in \mathbb R^u$ for channel $k \in [K]$. Here $[K]$ represents the integer set $\{ 1,2,\ldots, K\}$. The $i$th entry of CAM $I_{\mathrm{CAM}}(\mathbf x, c)$ is  given by 
{\small \begin{align}
    [I_{\mathrm{CAM}}(\mathbf x, c)]_i = (1/u) \sum_{k \in [K]} w_{k}^c A_{k, i}, ~ i \in [u], \label{eq: CAM}
\end{align}}
where $w_k^c$ is the linear classification weight that associates the channel $k$ with the class $c$, and $A_{k,i}$ denotes the $i$th element of $\mathbf A_k$. The rationale behind \eqref{eq: CAM} is that the classification score $f_c(\mathbf x)$ can be written as the average of CAM values \citep{zhou2016learning}, 
$f_c(\mathbf x ) = \sum_{i=1}^u [I_{\mathrm{CAM}}(\mathbf x, c)]_i$.
For visual explanation,  $I_{\mathrm{CAM}}(\mathbf x, c)$ is often up-sampled to the input dimension $d$ using bi-linear interpolation.
 
GradCAM \citep{selvaraju2017grad} generalizes CAM for CNNs without the architecture `global average pooling $\to$ softmax layer' over the final convolutional maps. 
Specifically, the weight $w_k^c$ in  \eqref{eq: CAM} is given by the gradient of the classification score $f_c(\mathbf x)$ with respect to (w.r.t.) the feature map $\mathbf A_k$,
$
    w_k^c = \frac{1}{u} \sum_{i=1}^u \frac{\partial f_c(\mathbf x)}{\partial A_{k,i}}
$.   GradCAM++  \citep{chattopadhay2018grad}, a generalized formulation of GradCAM, utilizes a more involved  weighted average
of the (positive) pixel-wise gradients but
provides a better localization map  if an image contains multiple occurrences of the
same class. In this work, we focus on CAM since it is computationally light and our models used in experiments  follow the architecture `global average pooling $\to$ softmax layer'.

 \begin{figure*}[htb]
  \centering
  \begin{adjustbox}{max width=1\textwidth }
  \begin{tabular}{@{\hskip 0.00in}c @{\hskip 0.00in}c | @{\hskip 0.00in} @{\hskip 0.02in} c @{\hskip 0.02in} | @{\hskip 0.02in} c @{\hskip 0.02in} |@{\hskip 0.02in} c @{\hskip 0.02in} 
  }
& 
\colorbox{lightgray}{ \textbf{Input image}}
&
\colorbox{lightgray}{ \textbf{CAM}} 
&  
\colorbox{lightgray}{ \textbf{GradCAM++}} 
&
\colorbox{lightgray}{  \textbf{IG}}

\\
 \begin{tabular}{@{}c@{}}  

\rotatebox{90}{\parbox{10em}{\centering  \textbf{Original example $\mathbf x$}}}
 \\

\rotatebox{90}{\parbox{10em}{\centering  \textbf{Adversarial example $\mathbf x^\prime$ 
}}}

\end{tabular} 
& 
\begin{tabular}{@{\hskip 0.02in}c@{\hskip 0.02in}}
\\
 \begin{tabular}{@{\hskip 0.02in}c@{\hskip 0.02in} }
 \parbox[c]{10em}{\includegraphics[width=10em]{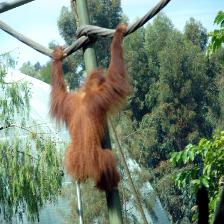}

 }  
\end{tabular}
 \\
 \begin{tabular}{@{\hskip 0.02in}c@{\hskip 0.02in}}
 \parbox[c]{10em}{\includegraphics[width=10em]{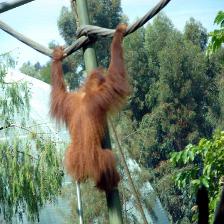}}  
\end{tabular}
\end{tabular}
&
\begin{tabular}{@{\hskip 0.02in}c@{\hskip 0.02in}}
 \begin{tabular}{@{\hskip 0.02in}c@{\hskip 0.02in}c@{\hskip 0.02in} }
      \begin{tabular}{@{\hskip 0.00in}c@{\hskip 0.00in}}
     \parbox{10em}{\centering   $I(\cdot, t)$}  
    \end{tabular} 
     &  
      \begin{tabular}{@{\hskip 0.00in}c@{\hskip 0.00in}}
       \parbox{10em}{\centering   $I(\cdot, y^\prime )$} 
    \end{tabular} 
\end{tabular}\\
 \begin{tabular}{@{\hskip 0.02in}c@{\hskip 0.02in}c@{\hskip 0.02in} }
 \parbox[c]{10em}{\includegraphics[width=10em]{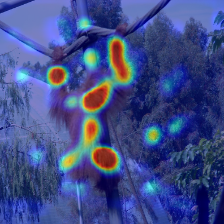}

 } &    \parbox[c]{10em}{\includegraphics[width=10em]{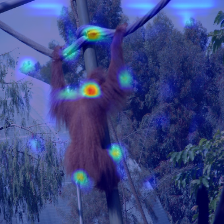}} 
\end{tabular}
 \\
 \begin{tabular}{@{\hskip 0.02in}c@{\hskip 0.02in}c@{\hskip 0.02in} }
 \parbox[c]{10em}{\includegraphics[width=10em]{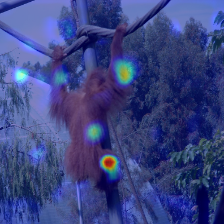}} &    \parbox[c]{10em}{\includegraphics[width=10em]{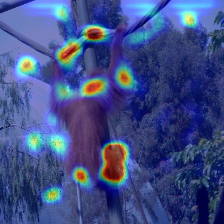}} \\
  \parbox{10em}{\centering  \large correlation: 0.4782}   &  \parbox{10em}{\centering \large correlation: 0.5039}  
\end{tabular}
\end{tabular}
&
\begin{tabular}{@{\hskip 0.02in}c@{\hskip 0.02in}}
 \begin{tabular}{@{\hskip 0.02in}c@{\hskip 0.02in}c@{\hskip 0.02in} }
      \begin{tabular}{@{\hskip 0.00in}c@{\hskip 0.00in}}
     \parbox{10em}{\centering   $I(\cdot, t)$}  
    \end{tabular} 
     &  
      \begin{tabular}{@{\hskip 0.00in}c@{\hskip 0.00in}}
       \parbox{10em}{\centering   $I(\cdot, y^\prime )$} 
    \end{tabular} 
\end{tabular}\\
 \begin{tabular}{@{\hskip 0.02in}c@{\hskip 0.02in}c@{\hskip 0.02in} }
 \parbox[c]{10em}{\includegraphics[width=10em]{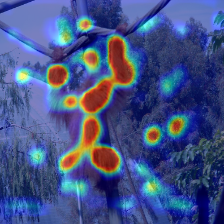}} &    \parbox[c]{10em}{\includegraphics[width=10em]{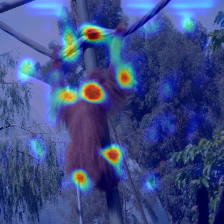}} 
\end{tabular}
 \\
 \begin{tabular}{@{\hskip 0.02in}c@{\hskip 0.02in}c@{\hskip 0.02in} }
 \parbox[c]{10em}{\includegraphics[width=10em]{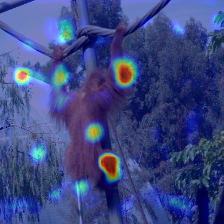}} &    \parbox[c]{10em}{\includegraphics[width=10em]{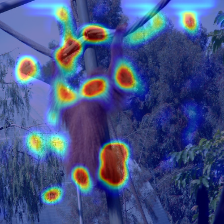}} 
 \\
  \parbox{10em}{\centering \large correlation: 0.5018}   &  \parbox{10em}{\centering  \large correlation: 0.5472}  
\end{tabular}
\end{tabular}
&
\begin{tabular}{@{\hskip 0.02in}c@{\hskip 0.02in}}
 \begin{tabular}{@{\hskip 0.02in}c@{\hskip 0.02in}c@{\hskip 0.02in} }
      \begin{tabular}{@{\hskip 0.00in}c@{\hskip 0.00in}}
     \parbox{10em}{\centering   $I(\cdot, t)$}  
    \end{tabular} 
     &  
      \begin{tabular}{@{\hskip 0.00in}c@{\hskip 0.00in}}
       \parbox{10em}{\centering   $I(\cdot, y^\prime )$} 
    \end{tabular} 
\end{tabular}\\
 \begin{tabular}{@{\hskip 0.02in}c@{\hskip 0.02in}c@{\hskip 0.02in} }
 \parbox[c]{10em}{\includegraphics[width=10em]{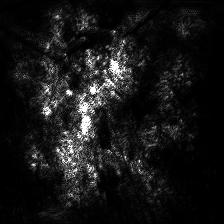}} &    \parbox[c]{10em}{\includegraphics[width=10em]{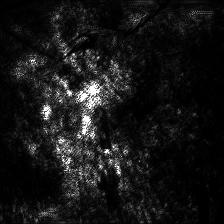}} 
\end{tabular}
 \\
 \begin{tabular}{@{\hskip 0.02in}c@{\hskip 0.02in}c@{\hskip 0.02in} }
 \parbox[c]{10em}{\includegraphics[width=10em]{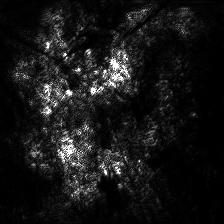}} &    \parbox[c]{10em}{\includegraphics[width=10em]{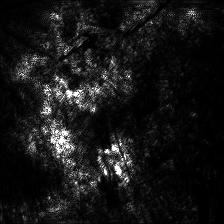}} 
 \\
  \parbox{10em}{\centering  \large correlation: 0.4040}   &  \parbox{10em}{\centering \large correlation: 0.3911}  
\end{tabular}
\end{tabular}

\end{tabular}
  \end{adjustbox}
\caption{\footnotesize{
Interpretation  ($I$) of benign  ($\mathbf x$) and adversarial ($\mathbf x^\prime$)    image from Restricted ImageNet   \citep{tsipras2018robustness} with respect to the true label {$y$=`monkey'} and the target label {$y^\prime$=`fish'}.
Here the adversarial example is generated by {10-step PGD attack with perturbation size 
$0.02$ on Wide-Resnet
\citep{madry2017towards},} and we consider three types of interpretation maps,  CAM, GradCAM++ and IG.  Given an interpretation method,  the first column is  $I(\mathbf x, y)$ versus $I(\mathbf x^\prime , y)$,   the second column is $I(\mathbf x, y^\prime)$ versus $I(\mathbf x^\prime , y^\prime)$, and  all maps under each category are normalized w.r.t. their largest value. At the bottom of each column, we quantify the resulting interpretation discrepancy by Kendall's Tau order rank correlation \citep{selvaraju2017grad} between every pair of $I(\mathbf x, i)$ and $I(\mathbf x^\prime, i)$ for $i = y$ or $y^\prime$. 
}}
\label{fig: inp_diff_adv_ori}
\end{figure*}

\paragraph{PSM-type methods.} 
PSM assigns   importance scores to individual pixels  toward explaining the classification decision about an input. Examples of commonly-used approaches include vanilla gradient \citep{simonyan2013deep}, guided backpropogation \citep{springenberg2014striving}, SmoothGrad \citep{smilkov2017smoothgrad}, and integrated gradient (IG) \citep{sundararajan2017axiomatic}. In particular, IG satisfies the \textit{completeness} attribution axiom that   PSM ought to obey.  
Specifically, it 
   averages gradient saliency maps for interpolations between an input $\mathbf x$ and a baseline image $\mathbf a$:
 {\small  \begin{align} \label{eq: IG}
 & [ I_{\mathrm{IG}}(\mathbf x, c)]_i  = (x_i -  a_i ) \int_{\alpha = 0}^{1} \frac{\partial f_c ( \mathbf a + \alpha (\mathbf x - \mathbf a) ) }{\partial x_i } d \alpha \nonumber \\
  & \approx (x_i -  a_i ) \sum_{j=1}^m \frac{\partial f_c ( \mathbf a + \frac{j}{m} (\mathbf x - \mathbf a) ) }{\partial x_i } \frac{1}{m},~ i\in [d], 
\end{align}}%
where $m$ is  the number of steps in the Riemman approximation
of the integral.
The \textit{completeness}   axiom \citep[Proposition\,1]{sundararajan2017axiomatic} states that $\sum_{i=1}^d  [ I_{\mathrm{IG}}(\mathbf x, c)]_i  = f_c(\mathbf x) - f_c(\mathbf a)$, where the baseline image $\mathbf a$ is often chosen such that $f_c(\mathbf a) \approx  0$, e.g., the black image. Note  that CAM also satisfies the \textit{completeness}   axiom. 
PSM is able to highlight fine-grained details in the image, but is computationally intensive and not quite
class-discriminative compared to CAM \citep{selvaraju2017grad}.

\paragraph{Interpretation discrepancy caused by adversarial perturbations.}
Let $\mathbf x^\prime = \mathbf x + \boldsymbol{\delta}$ represent an \textit{adversarial example} w.r.t. $\mathbf x$, where $\boldsymbol{\delta}$ denotes an \textit{adversarial perturbation}. 
By replacing the input image $\mathbf x$ with $\mathbf x^\prime$, a CNN will be fooled from the \textit{true label} $y$ to the  \textit{target (incorrect) label} $y^\prime$. 
It was recently shown in \citep{xu2018structured, xu2019interpreting} that the adversary could introduce an evident \textit{interpretation discrepancy} w.r.t. \textit{both} the true and the target label in terms of $I(\mathbf x, y)$ vs. $I(\mathbf x^\prime , y)$, and $I(\mathbf x, y^\prime )$ vs. $I(\mathbf x^\prime , y^\prime )$. {An illustrative example is provided in Figure\,\ref{fig: inp_diff_adv_ori}. We see that an adversary \textit{suppresses} the network interpretation w.r.t. the true label $y$ but \textit{promotes} the interpretation w.r.t. the target label $y^\prime$. We also observe that compared to IG, CAM and GradCAM++ better localize class-specific discriminative regions.}

The example in Figure\,\ref{fig: inp_diff_adv_ori} provides two implications   on   the  robustness of classification versus    interpretation discrepancy.
First, an adversarial example designed for misclassification gives rise to   interpretation discrepancy. Spurred by that,  the problem of {interpretability sneaking attack} arises  \citep{ZWJ+18,subramanya2018towards}: 
 One may wonder whether or not it is easy to generate adversarial examples that {mistake classification} but keep {interpretation  intact}.
If such adversarial vulnerability exists, it could have serious consequences when classification and interpretation are jointly used in  tasks like medical diagnosis ~\citep{subramanya2018towards}, and call into question the faithfulness of interpretation to network classification. It is also suggested from interpretation discrepancy that an interpreter itself could be quite sensitive to input perturbations (even if they were not designed for misclassification). Spurred by that, the robustness of interpretation provides a supplementary robustness metric for CNNs \citep{ghorbani2019interpretation, dombrowski2019explanations,chen2019robust}.

\section{Robustness of Classification vs. Robustness of Interpretation }\label{sec: attack}

In this section, we revisit the validity of  interpretability sneaking attack (ISA) from the perspective of interpretation discrepancy. We show that it is in fact quite challenging to force an adversarial example to mitigate its associated interpretation discrepancy. 
Further, we  
propose a novel measure of interpretation discrepancy, and theoretically show that constraining it prevents the   success of   adversarial attacks (for misclassification).

Previous work  \citep{ZWJ+18,subramanya2018towards} showed that  it is \textit{not difficult} to prevent adversarial examples from having large interpretation discrepancy  when   the  latter is measured w.r.t.  a \textit{single} class label (either the true label $y$ or the target  label $y^\prime$). However, we see from Figure\,\ref{fig: inp_diff_adv_ori} that the prediction-evasion adversarial attack  alters interpretation maps
w.r.t. \textit{both}  $y$ and $y^\prime$. This motivates us to rethink whether the single-class interpretation discrepancy measure is proper, and whether ISA is truly easy to bypass an interpretation discrepancy check.

We consider the   generic form of $\ell_p$-norm based interpretation discrepancy,
{\small\begin{align}\label{eq: inp_dis}
    \mathcal D \left ( \mathbf x, \mathbf x^\prime  \right ) =  (  {1}/{|\mathcal C|} )
    \sum_{i \in \mathcal C} \left \| 
  I(\mathbf x, i) - I(\mathbf x^\prime , i)  \right \|_p,
\end{align}}
where recall that $\mathbf x$ and $\mathbf x^\prime$ are natural and adversarial examples respectively, 
 $I$ represents an interpreter, e.g., CAM or IG,
$\mathcal C$ denotes the set of class labels used in  $I$, $| \mathcal C |$ is the cardinality of $\mathcal C$, and  we consider $p \in \{ 1,2\}$  in this paper. Clearly,  a specification of \eqref{eq: inp_dis} relies on  the choice of $\mathcal C$ and   $p$.
 The specification of \eqref{eq: inp_dis} with $\mathcal C = \{ y, y^\prime \}$ and $p = 1$ leads to the \textit{$\ell_1$ 2-class interpretation discrepancy measure,}
{\small\begin{align}\label{eq: inp_dis_2_l1}
    \mathcal D_{2,\ell_1} \left ( \mathbf x, \mathbf x^\prime  \right ) = (1/2) & \left ( \left \| 
  I(\mathbf x, y) - I(\mathbf x^\prime , y)  \right \|_1  \right. \nonumber \\
  & \left. + \left \| 
  I(\mathbf x, y^\prime ) - I(\mathbf x^\prime , y^\prime)  \right \|_1 \right ).
\end{align}}
\paragraph{Rationale behind  \eqref{eq: inp_dis_2_l1}.}
Compared to the previous works \citep{ZWJ+18,subramanya2018towards}  which used a single class label,
we choose $\mathcal C = \{ y, y^\prime \}$\footnote{In addition to the \textit{2-class} case, our experiments will also cover the \textit{all-class} case $\mathcal C = [C]$. }, motivated by the fact that an
interpretation discrepancy occurs w.r.t. both $y$ and $y^\prime$ (Figure\,\ref{fig: inp_diff_adv_ori}).
Moreover, although
Euclidean distance (namely, $\ell_2$ norm or its square)
is arguably one of the most commonly-used discrepancy metrics  \citep{ZWJ+18}, we show 
in Proposition\,\ref{prop: inp_measure} that 
the proposed interpretation discrepancy measure  $\mathcal D_{2,\ell_1} \left ( \mathbf x, \mathbf x^\prime  \right )$ has a perturbation-independent lower bound for
any \textit{successful} adversarial attack. This provides an explanation on why it could be difficult to mitigate the interpretation discrepancy caused by a successful  attack. As will be evident later, the use of $\ell_1$ norm also outperforms the $\ell_2$ norm in evaluation of interpretation discrepancy. 

\begin{prop}\label{prop: inp_measure}
Given a classifier $f(\x) \in \mathbb R^C$ and its interpreter  $I(\mathbf x, c)$ for $c \in [C]$, suppose that the interpreter satisfies the completeness axiom, namely, $\sum_i [ I(\mathbf x, c) ]_i =  f_c(\mathbf x)$ for a possible scaling factor $a$. For a natural example $\mathbf x$ and an adversarial example $\mathbf x^\prime  $ with prediction $y$ and $y^\prime$ ($\neq y$) respectively,  $\mathcal D_{2,\ell_1} \left ( \mathbf x, \mathbf x^\prime  \right )$ in \eqref{eq: inp_dis_2_l1} has the perturbation-independent lower bound,
{\small\begin{align} \label{eq: prop_inp_discrepancy}
      \mathcal D_{2,\ell_1} \left ( \mathbf x, \mathbf x^\prime  \right )  \geq (1/2) \left ( f_y(\x)  - f_{y'}(\x) \right ).
\end{align}}
\end{prop}
\textbf{Proof}: 
See proof and a generalization in Appendix\,\ref{app: proof_prop1}. \hfill $\square$

Proposition\,\ref{prop: inp_measure} connects
$\mathcal D_{2,\ell_1} \left ( \mathbf x, \mathbf x^\prime  \right ) $
with the classification margin   $ f_y(\x)  - f_{y'}(\x)$.  Thus, if a classifier has a large classification margin on the natural example $\mathbf x$, it will be difficult to find a \textit{successful}  adversarial attack with  \textit{small} interpretation discrepancy.  In other words, 
constraining the interpretation discrepancy 
prevents misclassification of a perturbed input since making its attack successful  becomes \textit{infeasible} under $\mathcal D_{2,\ell_1} \left ( \mathbf x, \mathbf x^\prime  \right )  <\frac{1}{2} \left ( f_y(\x)  - f_{y'}(\x) \right )$. Also, the completeness condition of $I$ suggests specifying \eqref{eq: inp_dis_2_l1} with
  CAM \eqref{eq: CAM} or IG \eqref{eq: IG}. Indeed, the robust attribution regularization proposed in \cite{chen2019robust} adopted  IG. 
In this paper, we focus on CAM due to its   light computation. In Appendix\,~\ref{app: proof_prop1}, we further extend Proposition\,\ref{prop: inp_measure} to interpreters satisfying a more general completeness axiom of the form $\sum_i [ I(\mathbf x, c) ]_i = g(f_c(\mathbf x))$, where $g$ is a monotonically increasing function.
In Appendix\,\ref{app: prop1_test}, we demonstrate the empirical tightness of \eqref{eq: prop_inp_discrepancy}.

\paragraph{{Attempt in generating ISA with minimum $\ell_1$ $2$-class interpretation discrepancy.}}

Next, we examine how the robustness of classification is coupled with the robustness of interpretation through the lens of   ISA. 
We pose the following optimization problem for design of ISA, which not only fools a classifier's decision but also   minimizes the resulting interpretation discrepancy,
{\small \begin{align}
    \hspace*{-0.1in}   \begin{array}{cl}
\displaystyle \minimize_{\boldsymbol{\delta}}         & \lambda \max \{ \max_{j \neq y^\prime } f_j(\mathbf x + \boldsymbol{\delta}) - f_{y^\prime }(\mathbf x + \boldsymbol{\delta}), -\tau  \} \\
& +  \mathcal D_{2,\ell_1} \left ( \mathbf x, \mathbf x + \boldsymbol{\delta} \right ) \\
    \st      & \| \boldsymbol{\delta } \|_\infty \leq \epsilon.
    \end{array}
    \hspace*{-0.1in}
    \label{eq: ISA}
\end{align}}%
In \eqref{eq: ISA},
the first term   corresponds to a C\&W-type attack loss \citep{carlini2017towards}, which reaches $-\tau$  if the attack succeeds in misclassification, 
 $\tau >  0$ (e.g., $0.1$ used in the paper) is a tolerance   on the classification margin of a successful attack      between the target label $y^\prime $  and the non-target top-1 prediction label,
$\mathcal D_{2, \ell_1}$ was defined by
\eqref{eq: inp_dis_2_l1}, $\lambda > 0$ is a regularization parameter that strikes a balance between the success of an attack and its resulting interpretation discrepancy, and $\epsilon > 0$ is a  (pixel-level) perturbation size.

To approach ISA \eqref{eq: ISA} with \textit{minimum} interpretation discrepancy, we perform a \textit{bisection}  on $\lambda$ until  there exists no successful attack that can be found when $\lambda$ further decreases.
We call an attack a \textit{successful ISA} if
the value of the attack loss  stays at $-\tau$ (namely, a valid adversarial example) and the minimum $\lambda$ is achieved (namely, the largest penalization on interpretation discrepancy). 
We solve problem \eqref{eq: ISA}   by projected gradient descent (PGD), with sub-gradients taken at non-differentiable points. We consider only targeted attacks to better evaluate the effect on interpretability of target classes, although this approach can be extended to an untargeted setting (e.g., by  target label-free interpretation discrepancy measure   introduced in the next section).

\begin{figure}[htb]
\centering
\begin{tabular}{cc} 
\hspace*{-0.1in} \includegraphics[width=0.227\textwidth]{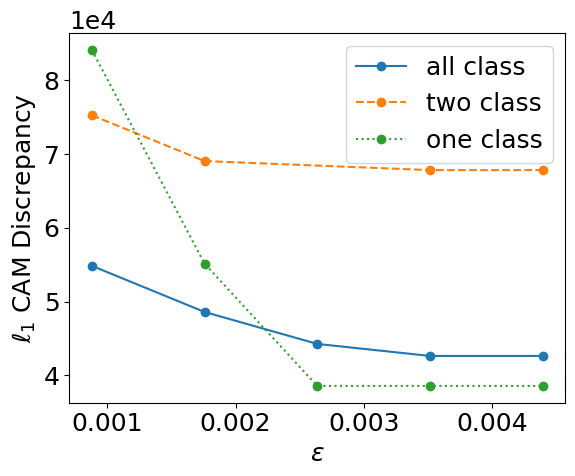} & \hspace*{-0.12in}
\includegraphics[width=0.235\textwidth]{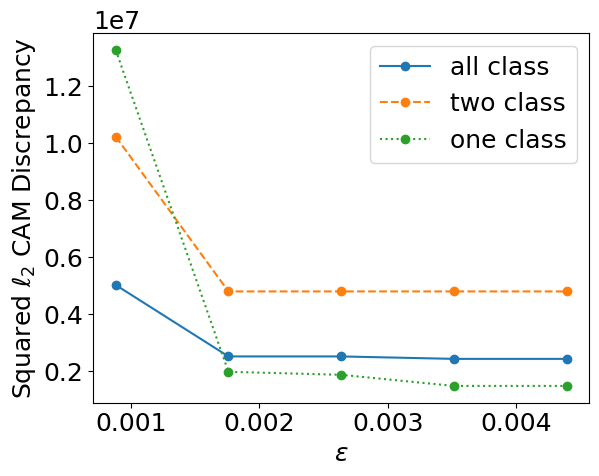}
\\
 {\footnotesize (a)} & \hspace*{-0.12in} {\footnotesize (b)} 
 \\
&  \hspace*{-1.8in}  \begin{adjustbox}{max width=0.49\textwidth}
  \begin{tabular}{@{\hskip 0.00in}c  @{\hskip 0.00in} @{\hskip 0.02in} c @{\hskip 0.02in} 
  }
 \begin{tabular}{@{}c@{}}  
\rotatebox{90}{\parbox{10em}{\centering  \textbf{ $I(\cdot, t)$ }}}
 \\

\rotatebox{90}{\parbox{10em}{\centering  \textbf{ $I(\cdot, y^\prime )$ }}}

\end{tabular} 
&
\begin{tabular}{@{\hskip 0.02in}c@{\hskip 0.02in}}
 \begin{tabular}{@{\hskip 0.02in}c@{\hskip 0.02in}c@{\hskip 0.02in}c@{\hskip 0.02in}c@{\hskip 0.02in}
 c@{\hskip 0.02in}
 }
      \begin{tabular}{@{\hskip 0.00in}c@{\hskip 0.00in}}
     \parbox{10em}{\centering   original image  $\mathbf x$}  
    \end{tabular} 
     &  
      \begin{tabular}{@{\hskip 0.00in}c@{\hskip 0.00in}}
       \parbox{10em}{\centering {10-step PGD attack 
       }  $\mathbf x^\prime$} 
    \end{tabular} 
     &  
      \begin{tabular}{@{\hskip 0.00in}c@{\hskip 0.00in}}
       \parbox{10em}{\centering    $\ell_1$ 2-class ISA $\mathbf x^\prime $ } 
    \end{tabular} 
     &  
      \begin{tabular}{@{\hskip 0.00in}c@{\hskip 0.00in}}
       \parbox{10em}{\centering   $\ell_1$ 1-class ISA $\mathbf x^\prime $ }
    \end{tabular} 
     &  
      \begin{tabular}{@{\hskip 0.00in}c@{\hskip 0.00in}}
       \parbox{10em}{\centering   $\ell_1$ all-class ISA $\mathbf x^\prime $ }
    \end{tabular} 

\end{tabular}\\
 \begin{tabular}{@{\hskip 0.02in}c@{\hskip 0.02in}c@{\hskip 0.02in}c@{\hskip 0.02in}c@{\hskip 0.02in}c@{\hskip 0.02in}
 }
 \parbox[c]{10em}{\includegraphics[width=10em]{figures/6_cam_true.png}} &    \parbox[c]{10em}{\includegraphics[width=10em]{figures/6_cam_true_adv_no_class.png}} 
 &    \parbox[c]{10em}{\includegraphics[width=10em]{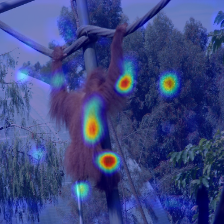}} 
 &    \parbox[c]{10em}{\includegraphics[width=10em]{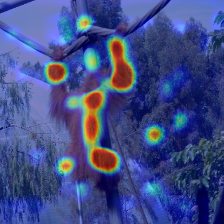}}
 &    \parbox[c]{10em}{\includegraphics[width=10em]{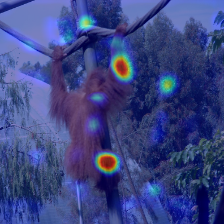}} 

 \\

  &     \parbox{10em}{\centering  \large correlation: 0.4782} 
 &     \parbox{10em}{\centering  \large correlation: 0.5213} 
 &     \parbox{10em}{\centering  \large correlation: 0.7107} 
 &     \parbox{10em}{\centering  \large correlation: 0.5342} 

\end{tabular}
  \\
 \begin{tabular}{@{\hskip 0.02in}c@{\hskip 0.02in}c@{\hskip 0.02in}c@{\hskip 0.02in}c@{\hskip 0.02in}c@{\hskip 0.02in}
 }
 \parbox[c]{10em}{\includegraphics[width=10em]{figures/6_cam_targ.png}} &    \parbox[c]{10em}{\includegraphics[width=10em]{figures/6_cam_targ_adv.png}} 
 &    \parbox[c]{10em}{\includegraphics[width=10em]{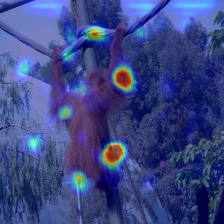}} 
 &    \parbox[c]{10em}{\includegraphics[width=10em]{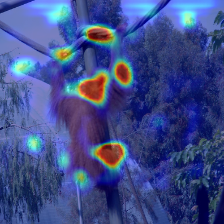}} 
 &    \parbox[c]{10em}{\includegraphics[width=10em]{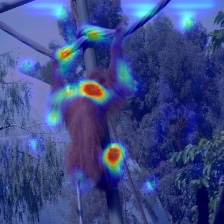}} 

 \\

  &     \parbox{10em}{\centering  \large correlation: 0.5039} 
 &     \parbox{10em}{\centering  \large correlation: 0.5416} 
 &     \parbox{10em}{\centering  \large correlation:  0.4129} 
 &     \parbox{10em}{\centering  \large correlation: 0.5561} 
\end{tabular}
\end{tabular}
\end{tabular}
  \end{adjustbox} 
  \\
  &  \hspace*{-1.8in}  {\footnotesize (c)}
\end{tabular}
\caption{  
\footnotesize{Interpretation  discrepancy induced by successful ISAs. Here the same  image as Figure\,\ref{fig: inp_diff_adv_ori} is considered.  (a) ISAs using CAM-based $\ell_1$ 1/2/all-class discrepancy measure versus perturbation size $\epsilon$, (b) ISAs using CAM-based  $\ell_2$ 1/2/all-class discrepancy measure versus $\epsilon$, (c) CAM interpretation of example in   Figure\,\ref{fig: inp_diff_adv_ori} and its adversarial counterparts from PGD attack  and ISAs. 
All interpretation maps  are normalized w.r.t. the common largest value.
At the bottom of each interpretation map  $I(\mathbf x^\prime, \cdot )$,    we quantify the   interpretation discrepancy by Kendall's Tau order rank correlation  between    $I(\mathbf x^\prime, i )$ and $I(\mathbf x, i )$ for $i \in \{ y, y^\prime\}$,
where $\mathbf x^\prime$ is obtained from PGD attack or each specification of ISA.
}}
\label{fig: exp_ISA}
\end{figure}

\paragraph{Successful ISA is accompanied by non-trivial $\ell_1$ 2-class  interpretation discrepancy.}
We then empirically justify that how the   choice of interpretation discrepancy measure 
plays a crucial role on drawing the relationship between robustness of classification and robustness of interpretation.
We generate successful ISAs   by solving problem  \eqref{eq: ISA} under different values of  the perturbation size $\epsilon$ and 
different specifications of the interpretation discrepancy measure \eqref{eq: inp_dis}, including $\ell_1$/$\ell_2$   1-class (true class $y$), $\ell_2$ 2-class, and 
 $\ell_1$/$\ell_2$ all-class measure. 
In  Figure\,\ref{fig: exp_ISA}-(a) and (b), we present the   interpretation discrepancy induced by successful ISAs versus the perturbation strength $\epsilon$. One may expect that a stronger ISA (with larger $\epsilon$) could more easily suppress the interpretation discrepancy. However, we observe that compared to $\ell_1$/$\ell_2$   1-class, $\ell_2$ 2-class, and $\ell_1$/$\ell_2$ all-class cases, it is quite difficult to mitigate the $\ell_1$ 2-class interpretation discrepancy  \eqref{eq: inp_dis_2_l1} even as the attack power goes up. This is verified by a)  its high interpretation discrepancy score and b) its flat slope of discrepancy score    against   $\epsilon$. 
 
Furthermore, Figure\,\ref{fig: exp_ISA}-(c)   shows CAMs of   adversarial examples w.r.t. the true label $y$ and the target label $y^\prime$ generated by $\ell_1$  1/2/all-class  ISAs. We observe that the 1-class measure 
could give a \textit{false} sense of ease of preventing   adversarial perturbations from interpretation discrepancy. Specifically,
although the interpretation discrepancy w.r.t. $y$  of the $\ell_1$ 1-class ISA  is minimized, the discrepancy w.r.t. $y^\prime$ remains large, supported by the observation that the resulting correlation between $I(\mathbf x^\prime, y^\prime)$ and $I(\mathbf x, y^\prime)$ is even smaller  than  that of PGD attack; see the $4$th column of  Figure\,\ref{fig: exp_ISA}-(c). Thus, the vulnerability of an image classifier (against adversarial perturbations) is accompanied by 
interpretation discrepancy only if the latter is properly measured. 
 We  refer readers to {Appendix\,\ref{app: NDS}} for more comprehensive experimental results on   the evaluation of interpretation discrepancy  through the lens of  ISA.

\section{Interpretability-Aware Robust Training}

We recall from Sec.\,\ref{sec: attack} that adversarial examples that intend  to fool a classifier could find it difficult to evade the $\ell_1$ 2-class interpretation discrepancy.  Thus,
 constraining the interpretation discrepancy  helps to prevent   misclassification. Spurred by that, we introduce an interpretability based defense method that penalizes interpretation discrepancy to achieve high classification robustness.

\paragraph{Target label-free   interpretation discrepancy.}
Different from attack generation, the $\ell_1$ 2-class   discrepancy measure  \eqref{eq: inp_dis_2_l1} cannot directly be used by a defender since  the target label $y^\prime $ specified by the adversary is \textit{not} known \textit{a priori}.
To circumvent this issue, we propose to approximate the interpretation discrepancy w.r.t. the target label by
weighting discrepancies from all non-true classes    
 according to their importance in prediction. This modifies  \eqref{eq: inp_dis_2_l1} to
{\small\begin{align}\label{eq: inp_dis_train}
\tilde {\mathcal D} \left ( \mathbf x, \mathbf x^\prime  \right ) = & (1/2)  \left \| 
  I(\mathbf x, y) - I(\mathbf x^\prime , y)  \right \|_1  \nonumber \\
&   \hspace*{-0.2in} 
  + (1/2)\sum_{i \neq t} \frac{e^{f(\x^\prime )_{i}}}{\sum_{i^{\prime}} e^{f(\x^\prime)_{i^{\prime}}}}||I(\x,i)-I(\x^\prime,i)||_1,
   \hspace*{-0.1in}
\end{align}}
where the softmax function $ \frac{e^{f(\x^\prime )_{i}}}{\sum_i e^{f(\x^\prime)_i}} $ adjusts the importance of non-true labels according to their classification confidence. Clearly, when  $\mathbf x^\prime$ succeeds in misclassification, 
the top-1 predicted class of $\mathbf x^\prime$ becomes the target label and the resulting interpretation discrepancy is most penalized.

\paragraph{Interpretability-aware robust training.}
We propose to train a classifier against the \textit{worst-case} interpretation discrepancy \eqref{eq: inp_dis_train}, yielding the min-max optimization problem
{\small\begin{align}
\hspace*{-0.1in}
\label{eq: adv_train_1}
\displaystyle \minimize_{\boldsymbol{\theta}} ~\mathbb E_{(\mathbf x, t)\sim \mathcal D_{\mathrm{train}}} \left [  f_{\mathrm{train}}(\boldsymbol{\theta};  \mathbf x, y)  +  \gamma \tilde{D}_{\mathrm{worst}}(\mathbf x, \mathbf x^\prime)
\right ],  
\hspace*{-0.1in}
\end{align}}%
where $\boldsymbol{\theta}$ denotes the model parameters to be learnt. In \eqref{eq: adv_train_1}, $ \mathcal D_{\mathrm{train}}$ denotes the training dataset, $f_{\mathrm{train}}$ is the training loss (e.g., cross-entropy loss),
$ \tilde{D}_{\mathrm{worst}}(\mathbf x, \mathbf x^\prime)$ denotes a measure of the {worst-case} interpretation discrepancy\footnote{For ease of notation we omit the dependence on      $\boldsymbol{\theta}$  in $\tilde {\mathcal D} \left ( \mathbf x, \mathbf x^\prime  \right )$} between the benign and the perturbed inputs $\mathbf x$ and $\mathbf x^{\prime}$, and the regularization parameter $\gamma > 0$ controls the tradeoff between clean accuracy and robustness of  network interpretability. 
Note that
 the commonly-used
 adversarial training method  \citep{madry2017towards} adopts  the \textit{adversarial loss} $\maximize_{  \| \boldsymbol{\delta }\|_\infty  \leq \epsilon }  f_{\mathrm{train}}(\boldsymbol{\theta}, \mathbf x + \boldsymbol{\delta};  \mathbf x, y) $ rather than 
 the \textit{standard training loss} in \eqref{eq: adv_train_1}. 
{Our experiments will show that the promotion of robust interpretation via \eqref{eq: adv_train_1} is able to achieve robustness in classification.}

Next, we introduce two types of worst-case interpretation discrepancy measure based on  our different views on input perturbations. That is, 
{\small\begin{align}
   &\tilde{D}_{\mathrm{worst}}(\mathbf x, \mathbf x^\prime) \Def   \maximize_{ \|  \boldsymbol{\delta}\|_\infty  \leq \epsilon }  \tilde {\mathcal D} \left ( \mathbf x, \mathbf x + \boldsymbol{\delta} \right ),   \label{eq: D_worst_inp}\\
   & \tilde{D}_{\mathrm{worst}}(\mathbf x, \mathbf x^\prime) \Def      \tilde {\mathcal D} \left ( \mathbf x, \mathbf x + \argmax_{  \| \boldsymbol{\delta} \|_\infty  \leq \epsilon } \,  [ f_{\mathrm{train}}(\boldsymbol{\theta};  \mathbf x + \boldsymbol{\delta}, y) ]  \right ),   \label{eq: D_worst_adv}
\end{align}}%
where $\tilde{\mathcal D}$ was defined in \eqref{eq: inp_dis_train}. 
In \eqref{eq: D_worst_inp} and \eqref{eq: D_worst_adv}, the input perturbation $\boldsymbol{\delta}$ represents the adversary shooting for misinterpretation and misclassification, respectively.
{
For ease of presentation, we call the proposed interpretability-aware robust training methods \textit{Int} and \textit{Int2} by using \eqref{eq: D_worst_inp} and \eqref{eq: D_worst_adv} in \eqref{eq: adv_train_1} respectively.
We will empirically show that both \textit{Int} and \textit{Int2} can achieve robustness in   classification and interpretation simultaneously. 
It is also worth noting that   \textit{Int2} training is conducted by alternative optimization: The inner maximization step w.r.t. $\boldsymbol{\delta}$ generates adversarial example  $\mathbf x^\prime$ for misclassification, and then forms $\tilde{D}_{\mathrm{worst}}(\mathbf x, \mathbf x^\prime)$; The outer minimization step minimizes the regularized standard training loss w.r.t. $\boldsymbol{\theta}$ by fixing $\mathbf x^\prime$, ignoring the dependence of $\mathbf x^\prime$ on $\boldsymbol{\theta}$.}

\paragraph{Difference from \citep{chen2019robust}.} The recent work \citep{chen2019robust}  proposed improving adversarial robustness  by leveraging  robust IG attributions. However, different from \citep{chen2019robust},
our approach is motivated by the importance of the $\ell_1$ \textit{2-class} interpretation discrepancy measure. We will show in Sec.\,\ref{sec: exp} that the incorporation of  interpretation discrepancy w.r.t. target class labels, namely, the second term in \eqref{eq: inp_dis_train}, plays an important role in boosting classification and interpretation robustness. 
We will also show that our proposed method is sufficient to improve adversarial robustness even in the absence of adversarial loss. This implies that
 robust interpretations alone helps robust classification   when interpretation maps are measured with a proper metric. 
 Furthermore,  we find that the robust attribution regularization method  \citep{chen2019robust} becomes less effective when the attack becomes stronger. Last but not least, beyond IG, 
 our proposed theory and method apply to any network interpretation method with the completeness axiom. 

\section{Experiments}\label{sec: exp}
In this section, we demonstrate the effectiveness of our proposed methods in $5$ aspects: a)  classification robustness against
PGD attacks  \citep{madry2017towards,athalye2018obfuscated},
b) defending against unforeseen adversarial attacks \citep{kang2019testing}, 
c)  computation efficiency, 
d)      interpretation robustness when facing attacks against interpretability  \citep{ghorbani2019interpretation},  and
e) visualization of perceptually-aligned robust features \citep{engstrom2019learning}. Our codes are available at \url{https://github.com/AkhilanB/Proper-Interpretability}

\paragraph{Datasets and CNN models.}
We evaluate networks trained on the MNIST and CIFAR-10 datasets, and a Restricted ImageNet (R-ImageNet) dataset used in \citep{tsipras2018robustness}.
We consider three models, \texttt{Small} (for MNIST and CIFAR), \texttt{Pool} (for MNIST) and \texttt{WResnet} (for CIFAR and R-ImageNet).
 \texttt{Small} is
  a small CNN architecture consisting of three convolutional layers of 16, 32 and 100 filters.
  \texttt{Pool} is
  a CNN architecture with two convolutional layers of 32 and 64 filters each followed by max-pooling which is adapted from \citep{madry2017towards}. \texttt{WResnet} is a Wide Resnet from \cite{zagoruyko2016wide} .

\paragraph{Attack models.}
{
 First, to evaluate robustness of classification, 
 we consider conventional
 \textit{PGD attacks} with different steps and perturbation sizes \citep{madry2017towards,athalye2018obfuscated} and \textit{unforeseen adversarial attacks}  \citep{kang2019testing} that are not used in robust training. 
 Second, to evaluate the robustness of interpretation, we consider \textit{attacks against interpretability (AAI)}  \citep{ghorbani2019interpretation, dombrowski2019explanations}, which  
produce  input perturbations to maximize the interpretation discrepancy rather than  misclassification.
We refer readers to {Appendix\,\ref{app: AAI}} for details on the generation of AAI.
Furthermore, we consider   \textit{ISA} \eqref{eq: ISA} under different   discrepancy measures to support our findings in Figure\,\ref{fig: exp_ISA}.  Details   are presented in  {Appendix\,\ref{app: NDS}}.
}

 \paragraph{Training methods.}
{
We consider \textit{$6$ baselines}: i) standard training (\textit{Normal}), ii) adversarial training (\textit{Adv}) \citep{madry2017towards}, iii) \textit{TRADES} \citep{zhang2019theoretically}, iv) \textit{IG-Norm} that uses  IG-based robust attribution regularization \citep{chen2019robust}, v) \textit{IG-Sum-Norm} (namely, IG-Norm with adversarial loss), and   
 vi) Int using $\ell_1$ $1$-class   discrepancy (\textit{Int-one-class}). 
Additionally, we consider \textit{$4$ variants} of our method: 
i) \textit{Int}, namely, \eqref{eq: adv_train_1} plus \eqref{eq: D_worst_inp}, ii) Int with adversarial loss (\textit{Int-Adv}), iii) \textit{Int2}, namely,\eqref{eq: adv_train_1} plus \eqref{eq: D_worst_adv},   and iv) Int2 with adversarial loss (\textit{Int2-Adv}).

{Unless specified otherwise, we choose the perturbation size    $\epsilon = 0.3$ on MNIST, $8/255$ on CIFAR and $0.003$ for R-ImageNet for robust training under an $\ell_\infty$ perturbation norm.} We refer readers to Appendix\,\ref{app: model} for more details.
Also, we  set the regularization parameter $\gamma$ as $0.01$  in  \eqref{eq: adv_train_1}; see a justification in 
Appendix\,\ref{app: gamma}.
Note that when training    \texttt{WResnet}, the IG-based robust training  methods (\textit{IG-Norm} and \textit{IG-Norm-Sum}) are excluded due to the prohibitive computation cost of computing IG. For our main experiments, we focus on the \texttt{Small} and \texttt{WResnet} architectures, but additional results on the \texttt{Pool} architecture are included in  Appendix\,\ref{app-tables}.

 }

\subsection{Classification   against prediction-evasion attacks}

\begin{figure}
     \centering
    \includegraphics[scale=0.45]{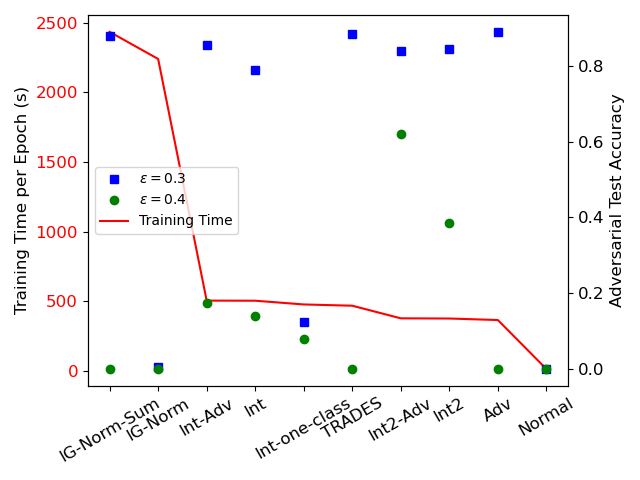}    
    \caption{
    \footnotesize{Computation time per epoch and adversarial test accuracy for a \texttt{Small} MNIST model trained with different methods.
    }}
    \label{fig:computation_accuracy}
\end{figure}

\paragraph{Robustness \& training efficiency.}
In  Figure \ref{fig:computation_accuracy}, we present the training time (left $y$-axis) and the adversarial test accuracy (right $y$-axis) for different training methods ($x$-axis) that are ranked in a decreasing order of computation complexity. Training times are {evaluated} on a 2.60 GHz Intel Xeon CPU.  
Here adversarial test accuracy (ATA) is found by performing $200$-step $\ell_\infty$-PGD  attacks of perturbation size  $\epsilon = 0.3$ and $0.4$ on the learned MNIST model \texttt{Small} over $200$ random test set points.
Note that all methods that use   adversarial losses (\textit{IG-Norm-Sum}, \textit{Int-Adv}, \textit{Int2-Adv}, \textit{TRADES} and \textit{Adv}) can yield robust classification at $\eps = 0.3$ (with ATA around $80\%$).
However, among  interpretability-regularized defense methods (\textit{IG-Norm}, \textit{Int-one-class}, \textit{Int}, \textit{Int2}), only the proposed \textit{Int} and \textit{Int2} methods provide competitive ATAs.  
As the PGD attack becomes stronger ($\epsilon = 0.4$),
\textit{Int} and \textit{Int2} based methods outperform all others in ATA. This implies  the benefit of  robust interpretation when facing  stronger prediction-evasion attacks; see more details in later results.

In Figure \ref{fig:computation_accuracy}, we also find that  both \textit{IG-Norm}
\citep{chen2019robust}  and \textit{Int-one-class} are insufficient to provide satisfactory ATA. The   verifies the importance on penalizing the \textit{2-class} interpretation discrepancy  to render   robust classification. We further observe that
IG-based methods  make training time (per epoch) significantly higher, e.g., $\geq 4$ times more than 
\textit{Int}. 

 \paragraph{Robustness against PGD attacks with different steps and perturbation sizes.}
It was shown in \citep{athalye2018obfuscated,carlini2019ami} that some defense methods cause 
\textit{obfuscated gradients}, which give a false sense of security. There exist two 
characteristic behaviors of obfuscated gradients: \textit{(a)} Increasing perturbation size does not increase attack success; \textit{(b)} One-step attacks perform better than iterative attacks. Motivated by that, we evaluate our interpretability-aware robust training methods under PGD attacks with different perturbation sizes and steps.

Table\,\ref{tab: pgd-acc-200} reports ATA of interpretability-aware robust training compared to various baselines over   MNIST and CIFAR, where $200$-step PGD attacks are conducted for robustness evaluation under different values of perturbation size $\epsilon$.
As we can see, ATA decreases as $\epsilon$ increases, violating the behavior (a)  of  obfuscated gradients. 
We also observe that compared to \textit{Adv} and \textit{TRADES}, \textit{Int} and \textit{Int2} achieve slightly worse  standard  accuracy ($\epsilon=0$) and
ATA on $\epsilon$ less than the value used for training. 
However,
when the $\epsilon$ used in the PGD attack achieves the value used for robust training,  \textit{Int} and \textit{Int2}  achieve better ATA than  \textit{Adv} on CIFAR-10 ($0.270$ and $0.290$ vs $0.170$). Interestingly, the advantage of \textit{Int} and \textit{Int2} becomes more evident as the adversary becomes stronger, i.e., $\epsilon > 0.3$ on MNIST and  $\epsilon > 8/255$ on CIFAR-10.
We highlight that such a robust classification is achieved by promoting robustness of interpretations alone (without using    adversarial loss). 

It is   worth mentioning that IG-Norm fails to defend PGD attack with $\epsilon = 0.3$ for the  MNIST model \texttt{Small}.  We further note that \textit{Int-one-class} performs much worse than \textit{Int}, supporting the \textit{importance of using a 2-class} discrepancy measure.
As will be evident later, \textit{IG-Norm} is also not the best to render robustness in interpretation (Table\,\ref{table: aai-topk-cam}). In Table\,\ref{tab:pgd-multistep} of Appendix\,\ref{app: vary_steps}, we further show that as the number of iterations of PGD attacks increases, the ATA of our proposed defensive schemes decreases accordingly. This violates the typical behavior (b) of obfuscated gradients.
\begin{table}[H]
\begin{center}
    \adjustbox{max width=0.48\textwidth}{
    \begin{tabular}[t]{lccccccc}
    \toprule
    Method & \multicolumn{1}{l}{$\eps=0$} & 0.05  & 0.1   & 0.2   & 0.3 & 0.35 & 0.4\\
    \midrule
    \multicolumn{8}{c}{MNIST, \texttt{Small}} \\
    \midrule
    Normal & \first{1.000} & 0.530 & 0.045 & 0.000 & 0.000 & 0.000 & 0.000\\
    Adv   & \third{0.980} & \second{0.960} & \second{0.940} & \second{0.925} & \first{0.890} & 0.010 & 0.000\\
    TRADES & 0.970 & \first{0.970} & \first{0.955} & \first{0.930} & \second{0.885} & 0.000 & 0.000\\
    IG-Norm & \second{0.985} & 0.950 & 0.895 & 0.410 & 0.005 & 0.000 & 0.000\\
    IG-Norm-Sum & 0.975 & \third{0.955} & \third{0.935} & \third{0.910} & 0.880 & 0.115 & 0.000 \\
    Int-one-class & 0.975 & 0.635 & 0.330 & 0.140 & 0.125 & 0.115 & 0.080 \\
    Int   &  0.950 & 0.930 & 0.905 & 0.840 & \highlight{0.790} & \highlight{0.180} & \highlight{0.140}\\
    Int-Adv &  0.935 & 0.945 & 0.905 & 0.880 & \highlight{0.855} & \highlight{0.355} & \highlight{0.175}\\
    Int2 & 0.950 & 0.945 & \third{0.935} & 0.890 & \highlight{0.845} & \highlight{0.555} & \highlight{0.385} \\
    Int2-Adv & 0.955 & 0.925 & 0.915 & 0.880 & \highlight{0.840} & \highlight{0.655} & \highlight{0.620} \\
    \midrule
          & \multicolumn{1}{l}{$\eps=0$} &
     \multicolumn{1}{l}{2/255} &\multicolumn{1}{l}{4/255} & \multicolumn{1}{l}{6/255} & \multicolumn{1}{l}{8/255} &  
     \multicolumn{1}{l}{9/255} &
     \multicolumn{1}{l}{10/255}\\
    \midrule
    \multicolumn{8}{c}{CIFAR-10, \texttt{WResnet}} \\
    \midrule
    Normal & \first{0.765} & 0.250 & 0.070 & 0.060 & 0.060 & 0.060 & 0.060\\
    Adv   & \third{0.720} & \third{0.605} & \second{0.485} & {0.330} & {0.170} & {0.145} & 0.085\\
    TRADES & \first{0.765} & \second{0.610} & 0.460 & 0.295 & 0.170 & 0.140 & {0.100} \\
    Int-one-class & 0.685 & 0.505 & 0.360 & 0.190 & 0.065 & 0.040 & 0.025\\
    Int   & \second{0.735} & \first{0.630} & \second{0.485} & \third{0.365} & \highlight{0.270} & \highlight{0.240} & \highlight{0.210}\\
    Int-Adv & 0.665 & 0.585 & \first{0.510} & \second{0.385} & \highlight{0.320} & \highlight{0.300} & \highlight{0.280}\\
    Int2 & 0.690 & 0.595 & 0.465 & 0.360 & \highlight{0.290} & \highlight{0.245} & \highlight{0.220} \\
    Int2-Adv & 0.680 & 0.585 & \second{0.485} & \first{0.405} & \highlight{0.335} & \highlight{0.310} & \highlight{0.285} \\
    \midrule
    \multicolumn{8}{c}{R-ImageNet, \texttt{WResnet}}   \\
    \midrule
    Normal & 0.770 & 0.070 & 0.035 & 0.030 & 0.040 & 0.030 & 0.030 \\
    Adv   & 0.790 & 0.455 & 0.230 & 0.100 & 0.070 & 0.060 & 0.050 \\
    Int   & 0.660 & 0.570 & 0.460 & 0.385 & \highlight{0.280} & \highlight{0.250} & \highlight{0.220} \\
    Int2  & 0.655 & 0.545 & 0.480 & 0.355 & \highlight{0.265} & \highlight{0.205} & \highlight{0.170} \\
    \bottomrule
    \end{tabular}
    }
    \caption{\footnotesize{Evaluation of 200-step PGD accuracy under different  perturbation sizes $\epsilon$. 
    ATA with $\epsilon = 0$ reduces to  standard test accuracy.}}
      \label{tab: pgd-acc-200}
\end{center}
\end{table}

\begin{table}[H]
  \centering
  \begin{adjustbox}{width=0.45\textwidth}
    \begin{tabular}{lccccc}
    \toprule
    Method 
    &
    \multicolumn{1}{l}{Gabor}
    & 
    \multicolumn{1}{l}{Snow} & \multicolumn{1}{l}{JPEG $\ell_\infty$} & \multicolumn{1}{l}{JPEG $\ell_2$} & \multicolumn{1}{l}{JPEG $\ell_1$} \\
    \midrule
    \multicolumn{6}{c}{CIFAR-10, \texttt{Small}} \\
    \midrule
    Normal 
    & 0.125 & 0.000 & 0.000 & 0.030 & 0.000 \\
    Adv   
    & 
    \second{0.190} & \third{0.115} & \second{0.460} & \highlight{0.380} & {0.230} \\
    TRADES 
    & \highlight{0.220} & 0.085 & 0.425 & 0.300 & 0.070 \\
    IG-Norm 
    & 0.155 & 0.015 & 0.000 & 0.000 & 0.000 \\
    IG-Norm-Sum 
    & \third{0.185} & {0.110} & \highlight{0.480} & \second{0.375} & 0.215 \\
    Int   
    & 0.160 & 0.105 & \third{0.440} & {0.345} & \second{0.260} \\
    Int-Adv
    & 0.150 & \second{0.120} & 0.340 & 0.310 & {0.235} \\
    Int2 & 0.130 & \third{0.115} & \third{0.440} & \third{0.365} & \highlight{0.295} \\
    Int2-Adv & 0.110 & \highlight{0.135} & 0.360 & 0.315 & \second{0.260} \\
    \bottomrule
    \end{tabular}
    \end{adjustbox}
    \caption{\footnotesize{ATA on different unforeseen   attacks in \citep{kang2019testing}. Best results in each column are \highlight{highlighted}.}}
  \label{tab: uar}
\end{table}

\paragraph{Robustness against unforeseen  attacks.}
In Table\,\ref{tab: uar}, we present ATA of interpretability-aware robust training and various baselines for defending  attacks (Gabor, Snow, JPEG $\ell_\infty$, JPEG $\ell_2$, and JPEG $\ell_1$) recently proposed in \citep{kang2019testing}. These attacks are called `unforeseen attacks' since they are not met by PGD-based robust training and often induce larger perturbations than conventional PGD attacks. We use the same attack parameters as used in \citep{kang2019testing} over 200 random test points. To compare with IG-based methods, we present results on the \texttt{Small} architecture since computing IG on the \texttt{WResnet} architecture is computationally costly. 
As we can see,
\textit{Int} and \textit{Int2} significantly outperform \textit{IG-Norm} especially under Snow and JPEG $\ell_p$ attacks. 
\textit{Int} and \textit{Int2} also yield competitive robustness compared to  the robust training methods that use the adversarial training loss  (\textit{Adv}, \textit{TRADES}, \textit{IG-Norm-Sum}, \textit{Int-Adv}, \textit{Int2-Adv}).

\subsection{Robustness of interpretation against AAI}
Recall that attack against interpretability (AAI) attempts to generate an adversarial interpretation map (namely, CAM in experiments) that is far away from the benign CAM of the original example w.r.t. the true label; see details in {Appendix\,\ref{app: AAI}}.  
The performance of AAI is then measured by the Kendall's Tau order rank correlation between the adversarial and the benign interpretation  maps     \citep{chen2019robust}. The higher the correlation is, the more robust  the model is in interpretation. Reported rank correlations are averaged over 200 random test set points.

In Table\,\ref{table: aai-topk-cam}, 
we present the performance of obtained robust models   against AAI with different attack strengths (in terms of the input perturbation size $\epsilon$);  see Table~\ref{tab: aai-topk-cam-full} of Appendix~\ref{app-tables} for results on additional dataset and networks.
The insights learned from Table\,\ref{table: aai-topk-cam} are summarized as below. First, the normally trained model (\textit{Normal}) does not  automatically offer robust interpretation, e.g.,  against AAI with   $\epsilon \geq 0.2$ in MNIST.
Second, the interpretation robustness of networks learned using  
adversarial training methods \textit{Adv} and \textit{TRADES} is worse than that learnt from interpretability-regularized training methods (except \textit{IG-Norm}) as 
the perturbation size $\epsilon$ increases ($\epsilon \geq 0.3$ for MNIST and $\epsilon \geq 8/255$ for R-ImageNet).
Third,  when     the adversarial training loss is not used,
our proposed methods \textit{Int} and \textit{Int2} are consistently more robust than \textit{IG-Norm}, and their advantage becomes more evident as $
\epsilon$ increases in MNIST.

\begin{table}[H]
\centering
\resizebox{0.47\textwidth}{!}{
\begin{tabular}[t]{lcccccc}
    \toprule
    Method & \multicolumn{1}{l}{$\eps=0.05$} & 0.1   & 0.2   & 0.3 & 0.35 & 0.4\\
    \midrule
    \multicolumn{7}{c}{MNIST, \texttt{Small}} \\
    \midrule
    Normal & 0.907 & 0.797 & 0.366 & -0.085 & -0.085 & -0.085\\
    Adv   & {0.978} & {0.955} & \third{0.910} & {0.857} & 0.467 & 0.136\\
    TRADES & {0.978} & {0.955} & 0.905 & 0.847 & 0.450 & 0.115 \\
    IG-Norm & 0.958 & 0.894 & 0.662 & 0.278 & 0.098 & 0.094\\
    IG-Norm-Sum & 0.976 & 0.951 & 0.901 & 0.850 & \second{0.659} & \second{0.389}\\
    Int-one-class & 0.874 & 0.818 & 0.754 & 0.692 & 
    0.461 & 0.278
    \\
    Int   & \first{0.982} & \second{0.968} & \second{0.941} & \highlight{0.913} & 
    \highlight{0.504} & \highlight{0.320}
    \\
    Int-Adv & {0.980} & \third{0.965} & {0.936} & \highlight{0.912} 
    &
    \highlight{0.527} & \highlight{0.348}
    \\
    Int2 & \first{0.982} & 0.967 & \second{0.941} & \highlight{0.918} & \highlight{0.612} & \highlight{0.351} \\
    Int2-Adv & \first{0.982} & \first{0.971} & \first{0.950} & \highlight{0.931} & \highlight{0.709} & \highlight{0.503} \\
    \midrule
          & \multicolumn{1}{l}{$\eps=2/255$} & \multicolumn{1}{l}{4/255} & \multicolumn{1}{l}{6/255} & \multicolumn{1}{l}{8/255} & 
          \multicolumn{1}{l}{9/255}
          & \multicolumn{1}{l}{10/255}
          \\
    \midrule
    \multicolumn{7}{c}{R-ImageNet, \texttt{WResnet}}  \\
    \midrule
    Normal & 0.851 & 0.761 & 0.705 & 0.673 & 0.659 & 0.619 \\
    Adv   & 0.975 & 0.947 & 0.916 & 0.884 & {0.870} & {0.858} \\
    Int   & 0.988 & 0.974 & 0.960 & \highlight{0.946} & \highlight{0.939} & \highlight{0.932} \\
    Int2  & 0.989 & 0.977 & 0.965 & \highlight{0.952} & \highlight{0.946} & \highlight{0.939} \\
    \bottomrule
    \end{tabular}
}
\caption{\footnotesize{Performance of AAI for different values of perturbation size $\epsilon$ in terms of   Kendall's  Tau  order  rank  correlation    between the original and adversarial interpretation maps. 
High interpretation robustness    corresponds  to large correlation value.}}
\label{table: aai-topk-cam}
\end{table}

\subsection{{Perceptually-aligned robust features}} 
\begin{figure}[htb]
  \centering
  \begin{adjustbox}{max width=0.41\textwidth }
  \begin{tabular}{@{\hskip 0.00in}c @{\hskip 0.01in} | @{\hskip 0.01in}c   @{\hskip 0.00in}   @{\hskip 0.00in} c @{\hskip 0.00in}   @{\hskip 0.00in} c @{\hskip 0.00in}  @{\hskip 0.0in} c @{\hskip 0.00in}    @{\hskip 0.0in} c
  }
\colorbox{lightgray}{ \textbf{\Large{Seed Images}}} 
&
\textcolor{blue}{ \textbf{\Large{Normal}}}
&

\textcolor{blue}{\Large{ \textbf{Adv}} }
& 

\textcolor{blue}{\Large{  \textbf{Int}}}
& 
\textcolor{blue}{ \Large{  \textbf{Int2}} }
\\
 \begin{tabular}{@{}c@{}}  
\\

 \begin{tabular}{@{\hskip 0.02in}c@{\hskip 0.02in} }
 \parbox[c]{10em}{\includegraphics[width=10em]{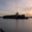}
 }  
\end{tabular} 
 \\
 \begin{tabular}{@{\hskip 0.02in}c@{\hskip 0.02in} }
 \parbox[c]{10em}{\includegraphics[width=10em]{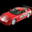}
 }  
\end{tabular}
 \\
 \begin{tabular}{@{\hskip 0.02in}c@{\hskip 0.02in} }
 \parbox[c]{10em}{\includegraphics[width=10em]{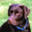}
 }  
\end{tabular}
 \\
 \begin{tabular}{@{\hskip 0.02in}c@{\hskip 0.02in} }
 \parbox[c]{10em}{\includegraphics[width=10em]{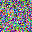}
 }  
\end{tabular}

\end{tabular} 
&
 \begin{tabular}{@{}c@{}}  
\\
 \begin{tabular}{@{\hskip 0.02in}c@{\hskip 0.02in} }
 \parbox[c]{10em}{\includegraphics[width=10em]{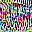}
 }  
\end{tabular} 
 \\
 \begin{tabular}{@{\hskip 0.02in}c@{\hskip 0.02in} }
 \parbox[c]{10em}{\includegraphics[width=10em]{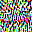}
 }  
\end{tabular}
 \\
 \begin{tabular}{@{\hskip 0.02in}c@{\hskip 0.02in} }
 \parbox[c]{10em}{\includegraphics[width=10em]{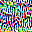}
 }  
\end{tabular}
 \\
  \begin{tabular}{@{\hskip 0.02in}c@{\hskip 0.02in} }
 \parbox[c]{10em}{\includegraphics[width=10em]{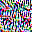}
 }  
\end{tabular}
 \\
 
\end{tabular}

&
 \begin{tabular}{@{}c@{}}  
\\

 \begin{tabular}{@{\hskip 0.02in}c@{\hskip 0.02in} }
 \parbox[c]{10em}{\includegraphics[width=10em]{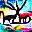}

 }  
\end{tabular} 
 \\
 \begin{tabular}{@{\hskip 0.02in}c@{\hskip 0.02in} }
 \parbox[c]{10em}{\includegraphics[width=10em]{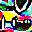}
 }  
\end{tabular}
 \\

 \begin{tabular}{@{\hskip 0.02in}c@{\hskip 0.02in} }
 \parbox[c]{10em}{\includegraphics[width=10em]{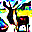}

 }  
\end{tabular}
 \\
 \begin{tabular}{@{\hskip 0.02in}c@{\hskip 0.02in} }
 \parbox[c]{10em}{\includegraphics[width=10em]{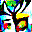}

 }  
\end{tabular}
 \\

\end{tabular} 
&
 \begin{tabular}{@{}c@{}}  
\\

 \begin{tabular}{@{\hskip 0.02in}c@{\hskip 0.02in} }
 \parbox[c]{10em}{\includegraphics[width=10em]{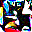}

 }  
\end{tabular} 
 \\

 \begin{tabular}{@{\hskip 0.02in}c@{\hskip 0.02in} }
 \parbox[c]{10em}{\includegraphics[width=10em]{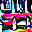}

 }  
\end{tabular}

 \\
 \begin{tabular}{@{\hskip 0.02in}c@{\hskip 0.02in} }
 \parbox[c]{10em}{\includegraphics[width=10em]{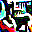}
 }  
\end{tabular}
 \\
 \begin{tabular}{@{\hskip 0.02in}c@{\hskip 0.02in} }
 \parbox[c]{10em}{\includegraphics[width=10em]{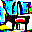}
 }  
\end{tabular}
 \\

\end{tabular} 
&
 \begin{tabular}{@{}c@{}}  
\\

 \begin{tabular}{@{\hskip 0.02in}c@{\hskip 0.02in} }
 \parbox[c]{10em}{\includegraphics[width=10em]{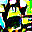}

 }  
\end{tabular} 
 \\
 \begin{tabular}{@{\hskip 0.02in}c@{\hskip 0.02in} }
 \parbox[c]{10em}{\includegraphics[width=10em]{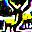}
 }  
\end{tabular}
 \\

 \begin{tabular}{@{\hskip 0.02in}c@{\hskip 0.02in} }
 \parbox[c]{10em}{\includegraphics[width=10em]{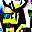}
 }  
\end{tabular}
\\
  \begin{tabular}{@{\hskip 0.02in}c@{\hskip 0.02in} }
 \parbox[c]{10em}{\includegraphics[width=10em]{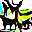}
 }  
\end{tabular}
 \\

\end{tabular} 

\end{tabular}
  \end{adjustbox}
\caption{\footnotesize{
Feature visualization at neuron $3$ under CIFAR-10  \texttt{WResnet} model trained by different methods. 
 Column $1$ contains  different seed images to maximize  neuron's activation. Columns $2$-$5$ contain  generated features w.r.t. each  seed image.
}}
\label{fig: robust_feature_cifar_wide_3}
\end{figure}

In Figure\,\ref{fig: robust_feature_cifar_wide_3}, we visually examine whether or not our proposed interpretability-aware training methods (\textit{Int} and \textit{Int2}) are able to render perceptually-aligned robust features similar to those found by \citep{engstrom2019learning} using \textit{Adv}. 
Figure\,\ref{fig: robust_feature_cifar_wide_3} shows that similar  texture-aligned robust features can   be acquired from networks trained using \textit{Int} and \textit{Int2} regardless of the choice of input seed image. This observation is consistent  with features learnt from \textit{Adv}. By contrast, the networks trained using \textit{Normal} and \textit{IG-Norm} fail to yield robust features; see results learnt from \textit{IG-Norm} under CIFAR-10 \texttt{Small} model in Appendix\,\ref{app: robust_feature}.

\section{Conclusion}
In this paper, we investigate the connection between network interpretability and adversarial robustness. We show theoretically and empirically that with the correct choice of discrepancy measure, it is difficult to hide adversarial examples from interpretation. We leverage this discrepancy measure to develop a interpretability-aware robust training method that displays 1) high classification robustness in a variety of settings and 2) high robustness of interpretation. Future work will extend our proposal to a  semi-supervised setting by incorporating unlabeled training data.

\section*{Acknowledgements}
This work was supported by the MIT-IBM Watson AI Lab. We particularly thank
 John M Cohn (MIT-IBM Watson AI Lab) for his generous help in computational resources. We also thank 
 David Cox (MIT-IBM Watson AI Lab) and Jeet Mohapatra
 (MIT) for their helpful discussions.


\bibliography{iclr2020_conference}
\bibliographystyle{iclr2020_conference}

\newpage
\clearpage
 \appendix

\setcounter{section}{0}
\setcounter{figure}{0}
\makeatletter 
\renewcommand{\thefigure}{A\@arabic\c@figure}
\makeatother
\setcounter{table}{0}
\renewcommand{\thetable}{A\arabic{table}}

\onecolumn

\section*{Appendix}

\section{Proof of Proposition\,\ref{prop: inp_measure}}\label{app: proof_prop1}
We first prove a generalization of Proposition\,\ref{prop: inp_measure} assuming a more general completeness axiom:  $\forall c \in [C],$ suppose $g(f_c(\mathbf x)) = \sum_i [ I(\mathbf x, c) ]_i $ where $g$ is a monotonically increasing function, e.g., a positive scaling function. The standard completeness axiom \citep{sundararajan2017axiomatic} uses the identity function for $g$: $g(z) = z$.

Using the generalized completeness axiom, we obtain that
\begin{align}
    g(f_{y'}(\x^\prime )) - g(f_{y'}(\x)) = & \sum_{i} ( [ I(\x^\prime,y') ]_i - [ I(\x,y') ]_i ) \nonumber \\
    \leq & \sum_{i} \left | [ I(\x^\prime,y') ]_i - [ I(\x,y') ]_i \right  | = \| I(\x^\prime,y') -  I(\x,y') \|_1. \label{eq: ineq_ft_prime}
\end{align}
Similarly, we have 
\begin{align}
    g(f_y(\x)) - g(f_y(\x^\prime)) \leq \| I(\x,y) - I(\x^\prime,y)\|_1. \label{eq: ineq_ft}
\end{align}
Adding \eqref{eq: ineq_ft_prime} and \eqref{eq: ineq_ft}   rearranging yields
\begin{align}
    [  g(f_{y'}(\x^\prime )) -g(f_y(\x^\prime))  ] + [   g(f_y(\x))  - g(f_{y'}(\x)) ] \leq \| I(\x^\prime,y') -  I(\x,y) \|_1 +  \| I(\x,y) - I(\x^\prime,y)\|_1.
\end{align}
Since $ f_{y'}(\x^\prime )  -f_y(\x^\prime) \geq 0$ and $g$ is monotonically increasing, $ g(f_{y'}(\x^\prime ))  -g(f_y(\x^\prime)) \geq 0$. We then have
$\| I(\x^\prime,y') -  I(\x,y') \|_1 +  \| I(\x,y) - I(\x^\prime,y)\|_1 \geq g(f_y(\x))  - g(f_{y'}(\x))$, which implies:
\begin{align}
    \mathcal{D}_{2,\ell_1}(\x,\x') \geq (1/2) [g(f_y(\x)) - g(f_{y'}(\x))].
\end{align}
This is a generalization of the bound in Proposition\,\ref{prop: inp_measure}. Taking $g(z) = z$ yields Equation~\eqref{eq: prop_inp_discrepancy}.

\mycomment{
For $\forall c \in [C]$, by the completeness axiom we have  $ f_c(\mathbf x) = \sum_i [ I(\mathbf x, c) ]_i $.  Using this fact, we obtain that
\begin{align}
    f_{y'}(\x^\prime ) - f_{y'}(\x) = & \sum_{i} ( [ I(\x^\prime,y') ]_i - [ I(\x,y') ]_i ) \nonumber \\
    \leq & \sum_{i} \left | [ I(\x^\prime,y') ]_i - [ I(\x,y') ]_i \right  | = \| I(\x^\prime,y') -  I(\x,y') \|_1. \label{eq: ineq_ft_prime}
\end{align}
Similarly, we have 
\begin{align}
    f_y(\x) - f_y(\x^\prime) \leq \| I(\x,y) - I(\x^\prime,y)\|_1. \label{eq: ineq_ft}
\end{align}
Adding \eqref{eq: ineq_ft_prime} and \eqref{eq: ineq_ft}   rearranging yields
\begin{align}
    [  f_{y'}(\x^\prime ) -f_y(\x^\prime)  ] + [   f_y(\x)  - f_{y'}(\x) ] \leq \| I(\x^\prime,y') -  I(\x,y') \|_1 +  \| I(\x,y) - I(\x^\prime,y)\|_1.
\end{align}
Since $ f_{y'}(\x^\prime )  -f_y(\x^\prime) \geq 0$, we then have
$\| I(\x^\prime,y') -  I(\x,y') \|_1 +  \| I(\x,y) - I(\x^\prime,y)\|_1 \geq f_y(\x)  - f_{y'}(\x)$, which is equivalent to \eqref{eq: prop_inp_discrepancy}.
}

\clearpage

\section{Interpretability Sneaking Attack (ISA): Evaluation and Results}\label{app: NDS}
In what follows, we provide additional experiment results on examining the relationship between classification robustness and interpretation robustness through the lens of ISA.
We evaluate the effect of interpretation discrepancy measure  on ease of finding ISAs. Spurred by Figure\,\ref{fig: exp_ISA},
such an effect is quantified by calculating minimum discrepancy required in generating ISAs against different values of  perturbation size $\epsilon$ in \eqref{eq: ISA}.
We conduct experiments over $4$ network interpretation methods: i) CAM, ii) GradCAM++, iii) IG, and iv)  internal representation  at the penultimate (pre-softmax) layer (denoted by \textit{Repr}).

In order to fairly compare  among different interpretation methods, we compute a \textit{normalized {discrepancy score (NDS)}} extended from \eqref{eq: inp_dis}:
$  \mathcal D_{\mathrm{norm}} =
\frac{1}{|\mathcal C|} \sum_{i \in \mathcal C} \left \| 
   \frac{    I(\mathbf x, i) -  I(\mathbf x^\prime, i)}{\max_{{j}} I(\mathbf x, i)_{{j}} - \min_{{j}} I(\mathbf x, i)_{{j}}} \right \|_p
$.
A larger value of NDS implies the more difficulty for   ISA to alleviate  interpretation discrepancy from adversarial perturbations.
To quantify the strength of ISA against  the  perturbation size $\epsilon$, 
we compute an additional quantity called \textit{normalized slope (NSL)} that measures the relative change of NDS   for $\epsilon \in [\check \epsilon, \hat \epsilon]$:
$\mathcal S_{\mathrm{norm}} = \frac{|\mathcal D_{\mathrm{norm}}^{(\hat \epsilon)} - \mathcal D_{\mathrm{norm}}^{(\check \epsilon)} |/\mathcal D_{\mathrm{norm}}^{(\check \epsilon)}}{(\hat \epsilon - \check \epsilon)/\check \epsilon}$.
The smaller NSL is, the more difficult it is for ISA to resist network interpretation changes as $\epsilon$ increases.  
In our experiment, we choose $\check \epsilon = \epsilon^*$ and $\hat \epsilon = 1.6 \, \epsilon^*$, where $\epsilon^*$ is the minimum perturbation size required for a successful PGD attack. Here we perform binary search over $\epsilon$ to find its smallest value for misclassification. Reported NDS and NSL statistics are averaged over a test set.

In Table\,\ref{tab:adv_train_sub}, we present   NDS and NSL of ISAs generated under  different realizations of interpretation discrepancy measure \eqref{eq: inp_dis}, each of which is given by a combination of interpretation method (CAM, GradCAM++, IG or Repr), $\ell_p$ norm ($p \in \{ 1,2\}$) and  number of interpreted classes. Note that Repr is independent of the number of classes, and thus we report NDS and NSL corresponding to Repr in the 2-class column of Table\,\ref{tab:adv_train_sub}.  
Given an $\ell_p$ norm and an interpretation method, we consistently find that the use of 2-class    measure 
achieves  the \textit{largest NDS} and \textit{smallest NSL} \textit{at the same time}. This implies that the 2-class  discrepancy measure
increases the difficulty for ISA to  evade a network interpretability check. Moreover, given a class number and an interpretation method, we see that NDS under $\ell_1$ norm is greater than that under $\ell_2$ norm, since the former is naturally an upper bound of the latter. Also, the use of $\ell_1$ norm often yields a smaller value of NSL, implying that the $\ell_1$-norm based discrepancy measure is more resistant to ISA. Furthermore, by fixing the combination of $\ell_1$ norm and $2$ classes,  we observe that  IG   
is the most resistant to ISA 
due to its relatively high NDS and low ISA, and  Repr yields the worst performance. 
However, compared to CAM,  the computation cost of IG increases dramatically as the input dimension,  the model size, and  the number of steps in   Riemman approximation increase. We find that it becomes infeasible to generate ISA using IG  for \texttt{WResnet}  under R-ImageNet within 200 hours.

\begin{table}[htb]
\centering
  \begin{adjustbox}{max width=0.85\textwidth }
\begin{tabular}{@{}c|c|ccc|ccc@{}}
\toprule
\multirow{2}{*}{ Dataset }
& \multirow{2}{*}{ \parbox{0.8in}{\centering  Interpretation\\method} }
 & \multicolumn{3}{c}{$\ell_1$ norm}
 & \multicolumn{3}{c}{$\ell_2$ norm}
 \\ 
\cmidrule(lr){3-5} \cmidrule(lr){6-8}
& & 1-class & 2-class & all-class & 1-class & 2-class & all-class
 \\\midrule
\multirow{4}{*}{MNIST} & CAM & 3.0723/0.0810 & 3.2672/0.0223 & 2.5289/0.0414 & 0.3061/0.1505 & 0.5654/0.0321 & 0.4320/0.0459 \\ 
 & GradCAM++  & 3.1264/0.0814 & 3.1867/0.0221 & 2.5394/0.0366 & 0.3308/0.1447 & 0.5531/0.0289 & 0.4392/0.0456 \\ 
 & IG & 6.3604/0.0330 & {6.7884/0.0233} & 4.3667/0.2314 & 0.4476/0.0082 & 0.5766/0.0064 & 0.2160/0.0337 \\ 
 & Repr & n/a & 2.3668/0.0404 & n/a & n/a & 0.4129/0.0429 & n/a \\ 
 \midrule
\multirow{4}{*}{CIFAR-10} & CAM & 1.9523/0.1450 & 2.5020/0.0496 & 1.7898/0.0774 & 0.1313/0.2369 & 0.3613/0.0668 & 0.2746/0.0809 \\ 
 & GradCAM++  & 1.9355/0.1439 & 2.4788/0.0513 & 1.8020/0.0745 & 0.1375/0.2346 & 0.3577/0.0676 & 0.2758/0.0769 \\ 
 & IG & 4.9499/0.0188 & 4.9794/0.0177 & 2.8541/0.1356 & 0.1230/0.0110 & 0.1309/0.0092 & 0.0878/0.0235 \\ 
 & Repr & n/a & 1.7049/0.0785 & n/a & n/a & 0.1288/0.0056 & n/a \\ 
  \midrule
\multirow{4}{*}{  R-ImageNet} & CAM & 49.286/0.1005 & 61.975/0.0331 & 49.877/0.0557 & 1.9373/0.1526 & 2.6036/0.0791 & 2.0935/0.0863 \\ 
 & GradCAM++  & 39.761/0.1028 & 50.303/0.0453 & 42.390/0.0552 & 1.9185/0.1609 & 2.5869/0.0891 & 2.1151/0.0896 \\  
 & Repr & n/a & 46.892/0.0657 & n/a & n/a & 2.0730/0.0781 & n/a \\ 
 \bottomrule
\end{tabular}
\end{adjustbox}
\caption{\footnotesize{NDS and NSL (format   given by NDS/NSL) of successful ISAs generated 
under  different specifications of interpretation discrepancy measure \eqref{eq: inp_dis} and datasets MNIST, CIFAR and R-ImageNet.
Here a discrepancy measure with   large NDS and   small NSL indicates  a non-trivial challenge for ISA to mitigate interpretation discrepancy.}}
\label{tab:adv_train_sub}
\end{table}

\clearpage

\section{Attack against Interpretability (AAI)}\label{app: AAI}

Different from ISA, AAI produces input perturbations to   
  maximize the interpretation discrepancy while keeping the classification decision intact.
  Thus, AAI provides a means to evaluate the adversarial robustness in interpretations. 
  Since $y = \argmax_i f_i(\mathbf x) = \argmax_i f_i(\mathbf x^\prime ) = y^\prime $ in AAI, the 2-class  interpretation discrepancy measure \eqref{eq: inp_dis_2_l1}  reduces to its 1-class version.
   The problem of generating AAI is then cast as
 {\small\begin{align}
    \begin{array}{cl}
\displaystyle \minimize_{\boldsymbol{\delta}}         & \lambda \max \{ \max_{j \neq y } f_j(\mathbf x + \boldsymbol{\delta}) - f_{y }(\mathbf x + \boldsymbol{\delta}), 0 \} -  \mathcal D_{1} \left ( \mathbf x, \mathbf x + \boldsymbol{\delta} \right ) \\
    \st      & \| \boldsymbol{\delta } \|_\infty \leq \epsilon,
    \end{array}\label{eq: AAI}
\end{align}}%
where the first term  is a hinge loss to enforce   $f_y(\mathbf x + \boldsymbol{\delta}) \geq \max_{j \neq y } f_j(\mathbf x + \boldsymbol{\delta})$, namely,  $ \argmax_i f_i(\mathbf x^\prime ) = y $ (unchanged prediction under input perturbations), and $\mathcal D_1$ denotes a 1-class interpretation discrepancy measure, e.g., $\mathcal D_{1,\ell_1}$ from \eqref{eq: inp_dis_2_l1}, or the top-$k$ pixel difference between interpretation maps    \citep{ghorbani2019interpretation}. Similar to  \eqref{eq: ISA}, the regularization parameter $\lambda$ in \eqref{eq: AAI} strikes a balance between stealthiness in classification and variation in interpretations. {Experiments in Sec.\,\ref{sec: exp}} will show that the state-of-the-art defense methods against adversarial examples do not necessarily preserve robustness in interpretations as $\epsilon$ increases, although the prediction is not altered. For evaluation, AAI are found over 200 random test set points. AAI are computed assuming an $\ell_\infty$ perturbation norm for different values of $\eps$ using 200 attack steps with a step size of $0.01$.

\clearpage

\section{Additional Experimental Details}\label{app: model}

\paragraph{Models}
The considered network models all have a global average pooling layer followed by a fully connected layer at the end of the network. For our \texttt{WResnet} model, we use a Wide Residual Network ~\cite{zagoruyko2016wide} of scale $\times 1$ consisting of (16, 16, 32, 64) filters in the residual units.

\paragraph{Robust Training}
During robust training of all baselines, 40 adversarial steps are used for MNIST, 10 steps for CIFAR and 7 steps for R-ImageNet. For finding perturbed inputs for robust training methods, a step size of $0.01$ is used for MNIST, $2/255$ for CIFAR and $0.1$ for R-ImageNet. To ensure stability of all training methods, the size of perturbation  is increased during training from $0$ to a final value of $0.3$ on MNIST, $8/255$ on CIFAR and $0.003$ on R-ImageNet. The perturbation size schedule for all three datasets consists of regular training ($\eps=0$) for a certain number of training steps (MNIST: $2000$, CIFAR: $5000$, R-ImageNet: $5000$) followed by a linear increase in the perturbation size until the end of training. This is done to maintain relatively high non-robust accuracy. A batch size of 50 is used for MNIST, 128 for CIFAR and 64 for R-ImageNet. On MNIST and CIFAR, these parameters are chosen to be consistent with the implementation in \cite{madry2017towards} including adversarial steps (MNIST: 40, CIFAR: 10), the step size (MNIST: 0.01, CIFAR: 2/255), the batch size (MNIST: 50, CIFAR: 128), and perturbation size (MNIST: 0.3, CIFAR: 8/255). 
MNIST networks are trained for 100 epochs, CIFAR networks are trained for 200 epochs, slightly fewer than the approximately 205 used in \cite{madry2017towards}, and R-ImageNet networks are trained for 35 epochs. For all methods, training is performed using Adam with an initial learning rate of 0.0001 for MNIST and 0.001 for CIFAR and R-ImageNet, with the learning decayed by $\times 1/10$ at training steps 40000 and 60000 for CIFAR and 8000 and 16000 for R-ImageNet. We note that some prior work including \cite{madry2017towards} uses momentum-based SGD instead.

For robust training of IG-based methods, to reduce the relative training time to other methods, we use 5 steps in our Riemann approximation of IG, which reduces computation time from the 10   steps used during training in \cite{chen2019robust}). In addition, we use a regularization parameter of 1 for IG-Norm and IG-Norm-Sum to maintain consistency between both methods. Other training parameters, including the number of epochs (100), the number of adversarial steps (40), the $\ell_\infty$ adversarial step size (0.01), the Adam optimizer learning rate (0.0001), the batch size (50) and the adversarial perturbation size (0.3) are the same as used by \cite{chen2019robust} on MNIST.

In our implementation of TRADES, we use a regularization parameter (multiplying the regularization term) of 1 on all datasets. Other training parameters are the same as used by \cite{zhang2019theoretically} including the number of adversarial training steps (MNIST: 40, CIFAR: 10), the perturbation size (MNIST: 0.3, CIFAR: 8/255) and the $\ell_\infty$ adversarial step size (MNIST: 0.01, CIFAR: 2/255).

\paragraph{Evaluations}
For PGD evaluation, we use a maximum of 200 steps for PGD attacks, increasing from the maximum of 20 steps used in \cite{madry2017towards}, since we found that accuracy can continue to drop until 200 attack steps. For top-$K$ AAI evaluations, we use a value of $K=8$ over all datasets, which we found to be suitable for CAM interpretation maps.

\clearpage
\section{Empirical Tightness of Proposition 1}
\label{app: prop1_test}
To evaluate the tightness of the bound in Proposition~\ref{prop: inp_measure}, we compute the values of the discrepancy (LHS) and classification margin (RHS) in Equation~\eqref{eq: prop_inp_discrepancy} on \texttt{Small} models trained on MNIST and CIFAR-10. To show the distributions of the values of discrepancy or classification margin over the test dataset, in each setting, we report deciles of these values (corresponding to the inverse cumulative distribution function evaluated at $10 \%, 20 \%, ...$). As observed in Table~\ref{tab:bound_tightness}, we find that the gap between discrepancy (rows 1 and 3) and classification margin (rows 2 and 4) is small, particularly compared to the variation in these quantities within each row. This indicates that the bound in Proposition 1 is quite tight.

\begin{table}[H]
  \centering
    \begin{tabular}{lrrrrrrrrr}
    \toprule
     & Decile = 0.1   & 0.2   & 0.3   & 0.4   & 0.5   & 0.6   & 0.7   & 0.8   & 0.9 \\
    \midrule
    \multicolumn{10}{c}{MNIST, \texttt{Small}} \\
    \midrule
    Discrepancy & 5.91  & 6.73  & 7.28  & 9.05  & 9.82  & 10.94 & 13.06 & 15.98 & 18.58 \\
    Classification Margin & 5.23  & 5.92  & 6.49  & 7.51  & 8.11  & 9.41  & 11.40 & 13.39 & 16.60 \\
    \midrule
    \multicolumn{10}{c}{CIFAR-10, \texttt{Small}} \\
    \midrule
    Discrepancy & 0.86  & 2.04  & 2.84  & 3.84  & 4.36  & 5.16  & 6.43  & 7.17  & 10.43 \\
    Classification Margin & 0.52  & 1.45  & 2.25  & 2.59  & 3.54  & 4.34  & 5.49  & 6.28  & 8.43 \\
    \bottomrule
    \end{tabular}%
    \caption{Deciles of discrepancy and classification margin reported over a test set. Quantities are reported for \texttt{Small} models trained on MNIST and CIFAR-10.}
  \label{tab:bound_tightness}%
\end{table}%

\section{Experiments on Regularization Parameter $\gamma$} \label{app: gamma}
\begin{figure}[htb]
    \centering
    \includegraphics[scale=0.5]{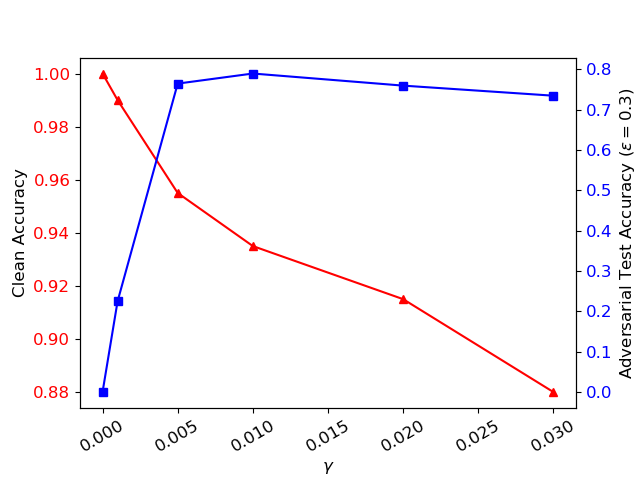}
    \caption{Clean test accuracy and adversarial test accuracy for a 
    \texttt{Small} MNIST model trained with \textit{Int} using different values of regularization parameter $\gamma$.}
    \label{fig:acc_ata_tradeoff}
\end{figure}

We conduct experiments for evaluating the sensitivity of the regularization parameter $\gamma$ in our proposed approach (namely, Int) under \texttt{Small} MNIST and CIFAR-10 models. For MNIST, adversarial test accuracy (ATA) and clean test accuracy results are plotted in Figure\,\ref{fig:acc_ata_tradeoff}. As illustrated, using different values of the hyperparameter $\gamma$ controls the tradeoff between clean accuracy and ATA, with smaller $\gamma$ yielding higher clean accuracy, but lower ATA (a value of $\gamma=0$ corresponds to normal training). We note that with the model tested, ATA stops increasing at a value of $\gamma=0.01$. Beyond this value, clean accuracy continues to decrease while ATA slightly decreases. These results indicate that by choosing an appropriate $\gamma$, it is possible to smoothly interpolate between normal training and maximally robust Int training. 
We also remark  that for all training $\eps$, ATA increases rapidly below $\gamma=0.01$, with a relatively small drop in clean accuracy. For instance, on CIFAR-10, at $\eps_{train} = 6/255$, when moving from $\gamma=0.005$ to $\gamma=0.01$, ATA increases by $13.0 \%$ with a drop of $5.5 \%$ in clean accuracy. We choose $\gamma = 0.01$ in our experiments.  

\clearpage

\section{Multi-step PGD Accuracy} \label{app: vary_steps}

Table\,\ref{tab:pgd-multistep} shows ATA of interpretability-aware robust training  against $k$-step PGD attacks, where $k \in \{ 1, 10, 100, 200\}$. As we can see, ATA decreases as $k$ increases. This   
 again verifies
  that the high robust accuracy obtained from our methods is not a result of obfuscated gradients. We  also see that  \textit{Int}
outperforms 
 \textit{IG-Norm} and \textit{Int-one-class} when facing stronger PGD attacks. Here the attack strength is characterized by
 the number of PGD steps.

\begin{table}[H] 
\vspace*{-0.05in}
\begin{center}
      \adjustbox{max width=0.5\textwidth}{
        \begin{tabular}[t]{lcccc}
        \toprule
        Method & \multicolumn{1}{l}{Steps$=1$} & 10    & 100   & 200 \\
        \midrule
        \multicolumn{5}{c}{MNIST, \texttt{Small}, $\eps=0.3$} \\
        \midrule
        Normal & \first{0.990} & 0.070 & 0.000 & 0.000 \\
        Adv   & \second{0.975} & \second{0.945} & \first{0.890} & \first{0.890} \\
        TRADES & \third{0.970} & \first{0.955} & \second{0.885} & \second{0.885} \\
        IG-Norm & \third{0.970} & 0.905 & 0.005 & 0.005 \\
        IG-Norm-Sum & \third{0.970} & \third{0.940} & \third{0.880} & \third{0.880} \\
                Int-one-class & 0.950 & 0.365 & 0.125 & 0.125 \\
        Int   & 0.935 & 0.910 & 0.790 & 0.790 \\
        Int-Adv & 0.950 & 0.905 & 0.855 & 0.855 \\
        Int2 & 0.950 & 0.935 & 0.845 & 0.845 \\
        Int2-Adv & 0.945 & 0.915 & 0.840 & 0.840 \\
        \midrule
        \multicolumn{5}{c}{CIFAR-10, \texttt{WResnet}, $\eps=8/255$} \\
        \midrule
        Normal & 0.470 & 0.075 & 0.060 & 0.060 \\
        Adv   & \second{0.590} & {0.205} & {0.185} & {0.185} \\
        TRADES & \second{0.590} & 0.180 & 0.165 & 0.165 \\
        Int-one-class & 0.505 & 0.100 & 0.060 & 0.060 \\
        Int   & \first{0.620} & {0.310} & {0.275} & {0.275} \\
        Int-Adv & {0.580} & \second{0.345} & \first{0.335} & \first{0.335} \\
        Int2 & \third{0.585} & \third{0.320} & \third{0.300} & \third{0.290} \\
        Int2-Adv & \third{0.585} & \first{0.360} & \first{0.335} & \first{0.335} \\
        \bottomrule
        \end{tabular}%
        }
        \caption{\footnotesize{Multi-step PGD accuracy.
        }}
        \label{tab:pgd-multistep}%
\end{center}
\end{table}

\clearpage
\newpage
\section{Additional Tables}
\label{app-tables}

\begin{table}[h!]
    \centering
    \begin{tabular}{lccccc}
    \toprule
    Method & \multicolumn{1}{l}{$\eps=0$} & 0.05  & 0.1   & 0.2   & 0.3 \\
    \midrule
    \multicolumn{6}{c}{MNIST, \texttt{Pool}} \\
    \midrule
    Normal & 0.990 & 0.435 & 0.070 & 0.000 & 0.000 \\
    Adv   & 0.930 & 0.885 & 0.835 & 0.695 & 0.535 \\
    TRADES & 0.955 & 0.910 & 0.870 & 0.720 & 0.455 \\
    IG-Norm & 0.980 & 0.940 & 0.660 & 0.050 & 0.000 \\
    IG-Norm-Sum & 0.920 & 0.885 & 0.840 & 0.700 & 0.540 \\
    Int-one-class & 0.975 & 0.885 & 0.720 & 0.200 & 0.130 \\
    Int   & 0.950 & 0.930 & 0.875 & 0.680 & 0.390 \\
    Int-Adv & 0.870 & 0.840 & 0.810 & 0.755 & 0.690 \\
    Int2 & 0.955 & 0.915 & 0.885 & 0.730 & 0.510 \\
    Int2-Adv & 0.865 & 0.830 & 0.805 & 0.760 & 0.705 \\
    \midrule
          & \multicolumn{1}{l}{$\eps=0$} & \multicolumn{1}{l}{2/255} & \multicolumn{1}{l}{4/255} & \multicolumn{1}{l}{6/255} & \multicolumn{1}{l}{8/255} \\
    \midrule
    \multicolumn{6}{c}{CIFAR-10, \texttt{Small}} \\
    \midrule
    Normal & 0.650 & 0.015 & 0.000 & 0.000 & 0.000 \\
    Adv   & 0.505 & 0.470 & 0.380 & 0.330 & 0.285 \\
    TRADES & 0.630 & 0.465 & 0.355 & 0.235 & 0.140 \\
    IG-Norm & 0.525 & 0.435 & 0.360 & 0.295 & 0.230 \\
    IG-Norm-Sum & 0.390 & 0.365 & 0.325 & 0.310 & 0.285 \\
    Int-one-class & 0.515 & 0.450 & 0.380 & 0.315 & 0.265 \\
    Int   & 0.530 & 0.450 & 0.345 & 0.290 & 0.215 \\
    Int-Adv & 0.675 & 0.145 & 0.005 & 0.000 & 0.000 \\
    Int2 & 0.470 & 0.430 & 0.360 & 0.330 & 0.260 \\
    Int2-Adv & 0.395 & 0.365 & 0.345 & 0.310 & 0.295 \\
    \bottomrule
    \end{tabular}
    \caption{200 steps PGD accuracy, additional results.}
    \label{tab: pgd-acc-200-full}
\end{table}

\begin{table}[h!]
    \centering
    \begin{tabular}{lccccc}
    \toprule
        Method & \multicolumn{1}{l}{$\eps=0$} & 0.05  & 0.1   & 0.2   & 0.3 \\
    \midrule
    \multicolumn{6}{c}{MNIST, \texttt{Pool}} \\
    \midrule
    Normal & 0.990 & 0.435 & 0.070 & 0.000 & 0.000 \\
    Adv   & 0.930 & 0.885 & 0.835 & 0.695 & 0.535 \\
    TRADES & 0.955 & 0.910 & 0.870 & 0.720 & 0.460 \\
    IG-Norm & 0.980 & 0.945 & 0.660 & 0.060 & 0.000 \\
    IG-Norm-Sum & 0.920 & 0.885 & 0.840 & 0.700 & 0.540 \\
    Int-one-class & 0.975 & 0.885 & 0.720 & 0.200 & 0.130 \\
    Int   & 0.950 & 0.930 & 0.875 & 0.680 & 0.385 \\
    Int-Adv & 0.870 & 0.840 & 0.810 & 0.755 & 0.700 \\
    Int2 & 0.955 & 0.915 & 0.885 & 0.730 & 0.510 \\
    Int2-Adv & 0.865 & 0.830 & 0.805 & 0.760 & 0.705 \\
    \midrule
          & \multicolumn{1}{l}{$\eps=0$} & \multicolumn{1}{l}{2/255} & \multicolumn{1}{l}{4/255} & \multicolumn{1}{l}{6/255} & \multicolumn{1}{l}{8/255} \\
    \midrule
    \multicolumn{6}{c}{CIFAR-10, \texttt{Small}} \\
    \midrule
    Normal & 0.650 & 0.015 & 0.000 & 0.000 & 0.000 \\
    Adv   & 0.505 & 0.470 & 0.380 & 0.330 & 0.285 \\
    TRADES & 0.630 & 0.465 & 0.355 & 0.235 & 0.140 \\
    IG-Norm & 0.525 & 0.435 & 0.360 & 0.295 & 0.230 \\
    IG-Norm-Sum & 0.390 & 0.365 & 0.325 & 0.310 & 0.285 \\
    Int-one-class & 0.515 & 0.450 & 0.380 & 0.315 & 0.265 \\
    Int   & 0.530 & 0.450 & 0.345 & 0.290 & 0.215 \\
    Int-Adv & 0.675 & 0.145 & 0.005 & 0.000 & 0.000 \\
    Int2 & 0.470 & 0.430 & 0.360 & 0.330 & 0.260 \\
    Int2-Adv & 0.395 & 0.365 & 0.345 & 0.310 & 0.295 \\
    \bottomrule
    \end{tabular}
    \caption{100 step PGD accuracy, additional results. }
\end{table}

\begin{table}[h!]
    \centering
    \begin{tabular}{lccccc}
    \toprule
        Method & \multicolumn{1}{l}{$\eps=0$} & 0.05  & 0.1   & 0.2   & 0.3 \\
    \midrule
    \multicolumn{6}{c}{MNIST, \texttt{Pool}} \\
    \midrule
    Normal & 0.990 & 0.470 & 0.135 & 0.135 & 0.135 \\
    Adv   & 0.930 & 0.885 & 0.845 & 0.845 & 0.845 \\
    TRADES & 0.955 & 0.910 & 0.870 & 0.870 & 0.870 \\
    IG-Norm & 0.980 & 0.945 & 0.705 & 0.705 & 0.705 \\
    IG-Norm-Sum & 0.920 & 0.885 & 0.850 & 0.850 & 0.850 \\
    Int-one-class & 0.975 & 0.885 & 0.750 & 0.750 & 0.750 \\
    Int   & 0.950 & 0.930 & 0.885 & 0.885 & 0.885 \\
    Int-Adv & 0.870 & 0.840 & 0.810 & 0.810 & 0.810 \\
    Int2 & 0.955 & 0.915 & 0.885 & 0.885 & 0.885 \\
    Int2-Adv & 0.865 & 0.830 & 0.805 & 0.805 & 0.805 \\
    \midrule
          & \multicolumn{1}{l}{$\eps=0$} & \multicolumn{1}{l}{2/255} & \multicolumn{1}{l}{4/255} & \multicolumn{1}{l}{6/255} & \multicolumn{1}{l}{8/255} \\
    \midrule
    \multicolumn{6}{c}{CIFAR-10, \texttt{Small}} \\
    \midrule
    Normal & 0.650 & 0.015 & 0.000 & 0.000 & 0.000 \\
    Adv   & 0.505 & 0.470 & 0.380 & 0.325 & 0.280 \\
    TRADES & 0.630 & 0.465 & 0.360 & 0.240 & 0.145 \\
    IG-Norm & 0.675 & 0.145 & 0.005 & 0.000 & 0.000 \\
    IG-Norm-Sum & 0.515 & 0.450 & 0.380 & 0.315 & 0.265 \\
    Int-one-class & 0.530 & 0.450 & 0.345 & 0.290 & 0.220 \\
    Int   & 0.525 & 0.435 & 0.360 & 0.295 & 0.235 \\
    Int-Adv & 0.390 & 0.365 & 0.325 & 0.310 & 0.285 \\
    Int2 & 0.470 & 0.430 & 0.360 & 0.330 & 0.265 \\
    Int2-Adv & 0.395 & 0.365 & 0.345 & 0.315 & 0.295 \\
    \bottomrule
    \end{tabular}
    \caption{10 step PGD accuracy, additional results.}
\end{table}

\begin{table}[h!]
    \centering
    \begin{tabular}{lcccc}
    \toprule
    Method & \multicolumn{1}{l}{$\eps=0.05$} & 0.1   & 0.2   & 0.3 \\
    \midrule
    \multicolumn{5}{c}{MNIST, \texttt{Pool}} \\
    \midrule
    Normal & 0.934 & 0.876 & 0.719 & 0.482 \\
    Adv   & 0.976 & 0.951 & 0.896 & 0.824 \\
    TRADES & 0.976 & 0.952 & 0.891 & 0.815 \\
    IG-Norm & 0.942 & 0.872 & 0.648 & 0.341 \\
    IG-Norm-Sum & 0.976 & 0.951 & 0.895 & 0.824 \\
    Int-one-class & 0.930 & 0.871 & 0.779 & 0.704 \\
    Int   & 0.964 & 0.928 & 0.852 & 0.771 \\
    Int-Adv & 0.977 & 0.957 & 0.921 & 0.891 \\
    Int2 & 0.969 & 0.941 & 0.885 & 0.832 \\
    Int2-Adv & 0.977 & 0.956 & 0.921 & 0.889 \\
    \midrule
          & \multicolumn{1}{l}{$\eps=2/255$} & \multicolumn{1}{l}{4/255} & \multicolumn{1}{l}{6/255} & \multicolumn{1}{l}{8/255} \\
    \midrule
    \multicolumn{5}{c}{CIFAR-10, \texttt{Small}} \\
    \midrule
    Normal & 0.694 & 0.350 & 0.116 & -0.031 \\
    Adv   & 0.958 & 0.907 & 0.849 & 0.783 \\
    TRADES & 0.940 & 0.867 & 0.781 & 0.689 \\
    IG-Norm & 0.810 & 0.552 & 0.308 & 0.131 \\
    IG-Norm-Sum & 0.958 & 0.907 & 0.847 & 0.779 \\
    Int-one-class & 0.961 & 0.918 & 0.871 & 0.820 \\
    Int   & 0.965 & 0.926 & 0.883 & 0.840 \\
    Int-Adv & 0.979 & 0.956 & 0.931 & 0.904 \\
    Int2 & 0.971 & 0.941 & 0.908 & 0.875 \\
    Int2-Adv & 0.980 & 0.959 & 0.938 & 0.914 \\
        \midrule
    \multicolumn{5}{c}{CIFAR-10, \texttt{WResnet}} \\
    \midrule
    Normal & 0.595 & 0.159 & 0.067 & -0.069\\
    Adv   & \second{0.912} & \second{0.816} & {0.724} & 0.629 \\
    TRADES & \first{0.918} & \first{0.832} & \second{0.747} & {0.652}  \\
    Int   & 0.859 & 0.763 & \third{0.746} & {0.682} \\
    Int-Adv & {0.885} & \third{0.803} & \first{0.751} & {0.696} \\
    Int2 & 0.868 & 0.779 & 0.708 & {0.674}  \\
    Int2-Adv & \third{0.889} & 0.788 & 0.721 & {0.672}  \\
    \bottomrule
    \end{tabular}
    \caption{Kendall rank correlation coefficients of Top-$k$ CAM attacks against interpretability found using 200 steps of PGD, additional results.
    }
    \label{tab: aai-topk-cam-full}
\end{table}

\clearpage
\newpage

\section{Additional Results on Robust Features}
\label{app: robust_feature}

 \begin{figure}[htb]
  \centering
  \begin{adjustbox}{max width=0.65\textwidth }
  \begin{tabular}{@{\hskip 0.00in}c @{\hskip 0.01in} | @{\hskip 0.01in}c   @{\hskip 0.00in}   @{\hskip 0.00in} c @{\hskip 0.00in}   @{\hskip 0.00in} c @{\hskip 0.00in}  @{\hskip 0.0in} c @{\hskip 0.00in}  @{\hskip 0.0in} c  
  }
\colorbox{lightgray}{ \textbf{\Large{Seed Images}}} 
&
\textcolor{blue}{ \textbf{\Large{Normal}}}
&
\textcolor{blue}{ \textbf{\Large{IG-Norm}}} 
&  

\textcolor{blue}{\Large{ \textbf{Adv}} }
& 

\textcolor{blue}{\Large{  \textbf{Int}}}
& 
\textcolor{blue}{ \Large{  \textbf{Int2}} }

\\
 \begin{tabular}{@{}c@{}}  
\vspace*{-0.1in}\\

 \begin{tabular}{@{\hskip 0.02in}c@{\hskip 0.02in} }
 \parbox[c]{10em}{\includegraphics[width=10em]{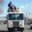}

 }  
\end{tabular} 
 \\

 \begin{tabular}{@{\hskip 0.02in}c@{\hskip 0.02in} }
 \parbox[c]{10em}{\includegraphics[width=10em]{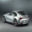}

 }  
\end{tabular}
 \\

 \begin{tabular}{@{\hskip 0.02in}c@{\hskip 0.02in} }
 \parbox[c]{10em}{\includegraphics[width=10em]{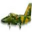}

 }  
\end{tabular}
 \\

 \begin{tabular}{@{\hskip 0.02in}c@{\hskip 0.02in} }
 \parbox[c]{10em}{\includegraphics[width=10em]{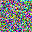}
 }  
\end{tabular}
\end{tabular} 
&
 \begin{tabular}{@{}c@{}}  
\vspace*{-0.1in}\\

 \begin{tabular}{@{\hskip 0.02in}c@{\hskip 0.02in} }
 \parbox[c]{10em}{\includegraphics[width=10em]{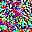}

 }  
\end{tabular}
 \\

 \begin{tabular}{@{\hskip 0.02in}c@{\hskip 0.02in} }
 \parbox[c]{10em}{\includegraphics[width=10em]{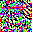}
 }  
\end{tabular}
 \\
 \begin{tabular}{@{\hskip 0.02in}c@{\hskip 0.02in} }
 \parbox[c]{10em}{\includegraphics[width=10em]{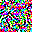}
 }  
\end{tabular}
 \\
 \begin{tabular}{@{\hskip 0.02in}c@{\hskip 0.02in} }
 \parbox[c]{10em}{\includegraphics[width=10em]{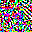}
 }  
\end{tabular}
\end{tabular} 
&
 \begin{tabular}{@{}c@{}}  
\vspace*{-0.1in}\\
 \begin{tabular}{@{\hskip 0.02in}c@{\hskip 0.02in} }
 \parbox[c]{10em}{\includegraphics[width=10em]{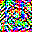}
 }  
\end{tabular} 
 \\
 \begin{tabular}{@{\hskip 0.02in}c@{\hskip 0.02in} }
 \parbox[c]{10em}{\includegraphics[width=10em]{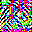}
 }  
\end{tabular}
 \\
 \begin{tabular}{@{\hskip 0.02in}c@{\hskip 0.02in} }
 \parbox[c]{10em}{\includegraphics[width=10em]{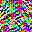}
 }  
\end{tabular}
 \\
 \begin{tabular}{@{\hskip 0.02in}c@{\hskip 0.02in} }
 \parbox[c]{10em}{\includegraphics[width=10em]{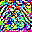}
 }  
\end{tabular}
\end{tabular} 
&

 \begin{tabular}{@{}c@{}}  
\vspace*{-0.1in}\\

\begin{tabular}{@{\hskip 0.02in}c@{\hskip 0.02in} }
\parbox[c]{10em}{\includegraphics[width=10em]{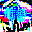}
 }  
\end{tabular}
 \\

 \begin{tabular}{@{\hskip 0.02in}c@{\hskip 0.02in} }
 \parbox[c]{10em}{\includegraphics[width=10em]{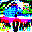}
 }  
\end{tabular}
 \\
 \begin{tabular}{@{\hskip 0.02in}c@{\hskip 0.02in} }
 \parbox[c]{10em}{\includegraphics[width=10em]{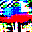}
 }  
\end{tabular}
 \\
 \begin{tabular}{@{\hskip 0.02in}c@{\hskip 0.02in} }
 \parbox[c]{10em}{\includegraphics[width=10em]{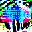}
 }  
\end{tabular}
\end{tabular} 
&
 \begin{tabular}{@{}c@{}}  
\vspace*{-0.1in}\\

 \begin{tabular}{@{\hskip 0.02in}c@{\hskip 0.02in} }
 \parbox[c]{10em}{\includegraphics[width=10em]{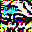}
 }  
\end{tabular} 
 \\

 \begin{tabular}{@{\hskip 0.02in}c@{\hskip 0.02in} }
 \parbox[c]{10em}{\includegraphics[width=10em]{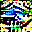}
 }  
\end{tabular}
 \\

 \begin{tabular}{@{\hskip 0.02in}c@{\hskip 0.02in} }
 \parbox[c]{10em}{\includegraphics[width=10em]{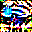}
 }  
\end{tabular}
 \\

 \begin{tabular}{@{\hskip 0.02in}c@{\hskip 0.02in} }
 \parbox[c]{10em}{\includegraphics[width=10em]{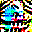}

 }  
\end{tabular}
\end{tabular} 
&
 \begin{tabular}{@{}c@{}}  
\vspace*{-0.1in}\\

 \begin{tabular}{@{\hskip 0.02in}c@{\hskip 0.02in} }
 \parbox[c]{10em}{\includegraphics[width=10em]{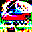}
 }  
\end{tabular}
 \\

 \begin{tabular}{@{\hskip 0.02in}c@{\hskip 0.02in} }
 \parbox[c]{10em}{\includegraphics[width=10em]{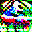}
 }  
\end{tabular}
 \\

 \begin{tabular}{@{\hskip 0.02in}c@{\hskip 0.02in} }
 \parbox[c]{10em}{\includegraphics[width=10em]{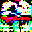}
 }  
\end{tabular}
 \\

 \begin{tabular}{@{\hskip 0.02in}c@{\hskip 0.02in} }
 \parbox[c]{10em}{\includegraphics[width=10em]{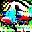}
 }  
\end{tabular}
\end{tabular} 

\end{tabular}
  \end{adjustbox}
\caption{\footnotesize{
Feature visualization at neuron $28$  under CIFAR-10  \texttt{Small} model trained by different methods. 
}}
\label{fig: robust_feature_cifar_10_28}
\end{figure}

\begin{figure}[htb]
  \centering
  \begin{adjustbox}{max width=0.65\textwidth }
  \begin{tabular}{@{\hskip 0.00in}c @{\hskip 0.01in} | @{\hskip 0.01in}c   @{\hskip 0.00in}   @{\hskip 0.00in} c @{\hskip 0.00in}   @{\hskip 0.00in} c @{\hskip 0.00in}  @{\hskip 0.0in} c @{\hskip 0.00in}    @{\hskip 0.0in} c 
  }
\colorbox{lightgray}{ \textbf{\Large{Seed Images}}} 
&
\textcolor{blue}{ \textbf{\Large{Normal}}}
&
\textcolor{blue}{ \textbf{\Large{IG-Norm}}} 

&  
\textcolor{blue}{\Large{ \textbf{Adv}} }
& 
\textcolor{blue}{\Large{  \textbf{Int}}}
& 
\textcolor{blue}{ \Large{  \textbf{Int2}} }
\\
 \begin{tabular}{@{}c@{}}  
\vspace*{-0.1in}\\

 \begin{tabular}{@{\hskip 0.02in}c@{\hskip 0.02in} }
 \parbox[c]{10em}{\includegraphics[width=10em]{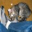}
 }  
\end{tabular} 
 \\
 \begin{tabular}{@{\hskip 0.02in}c@{\hskip 0.02in} }
 \parbox[c]{10em}{\includegraphics[width=10em]{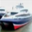}
 }  
\end{tabular}
 \\
 \begin{tabular}{@{\hskip 0.02in}c@{\hskip 0.02in} }
 \parbox[c]{10em}{\includegraphics[width=10em]{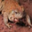}
 }  
\end{tabular}
 \\
 \begin{tabular}{@{\hskip 0.02in}c@{\hskip 0.02in} }
 \parbox[c]{10em}{\includegraphics[width=10em]{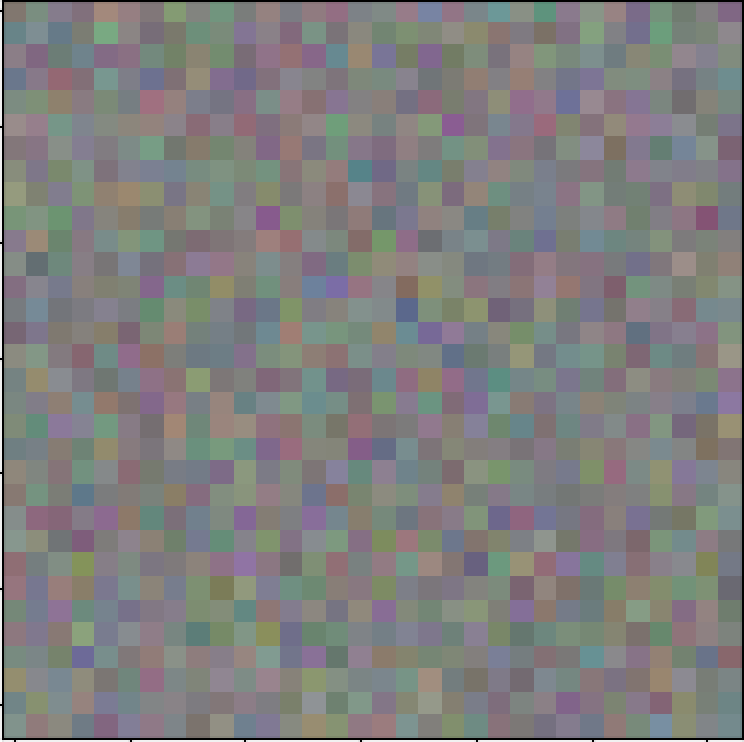}
 }  
\end{tabular}

\end{tabular} 
&
 \begin{tabular}{@{}c@{}}  
\vspace*{-0.1in}\\
 \begin{tabular}{@{\hskip 0.02in}c@{\hskip 0.02in} }
 \parbox[c]{10em}{\includegraphics[width=10em]{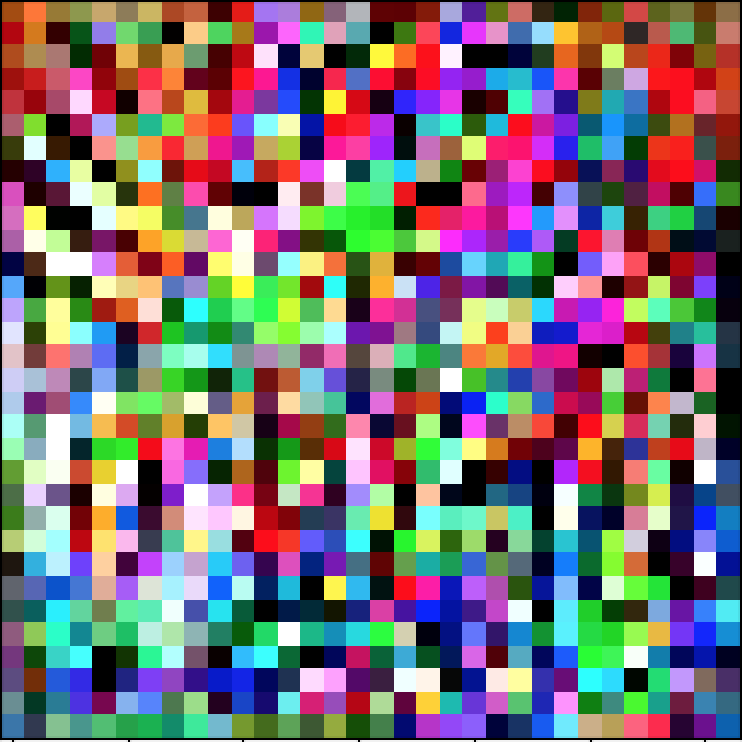}
 }  
\end{tabular} 
 \\
 \begin{tabular}{@{\hskip 0.02in}c@{\hskip 0.02in} }
 \parbox[c]{10em}{\includegraphics[width=10em]{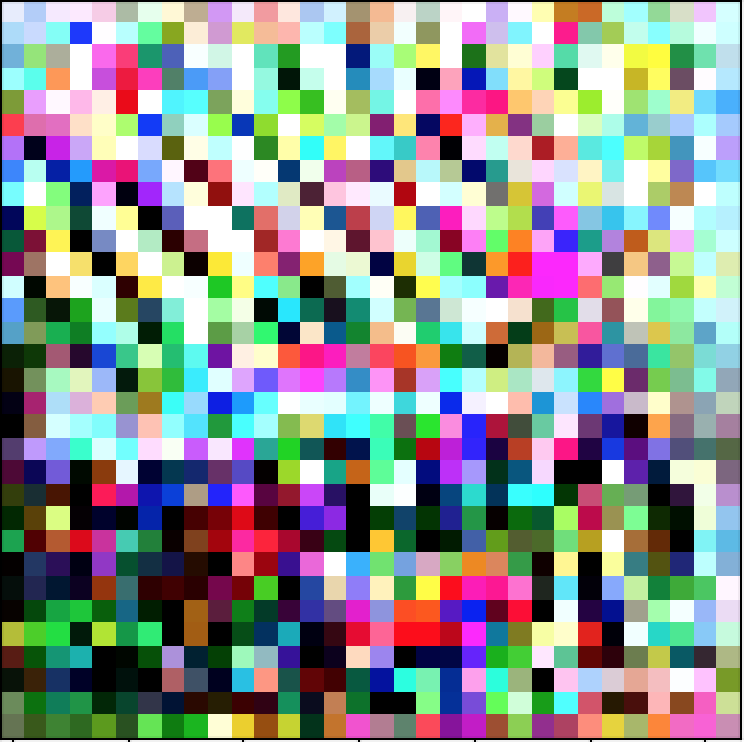}
 }  
\end{tabular}
 \\
 \begin{tabular}{@{\hskip 0.02in}c@{\hskip 0.02in} }
 \parbox[c]{10em}{\includegraphics[width=10em]{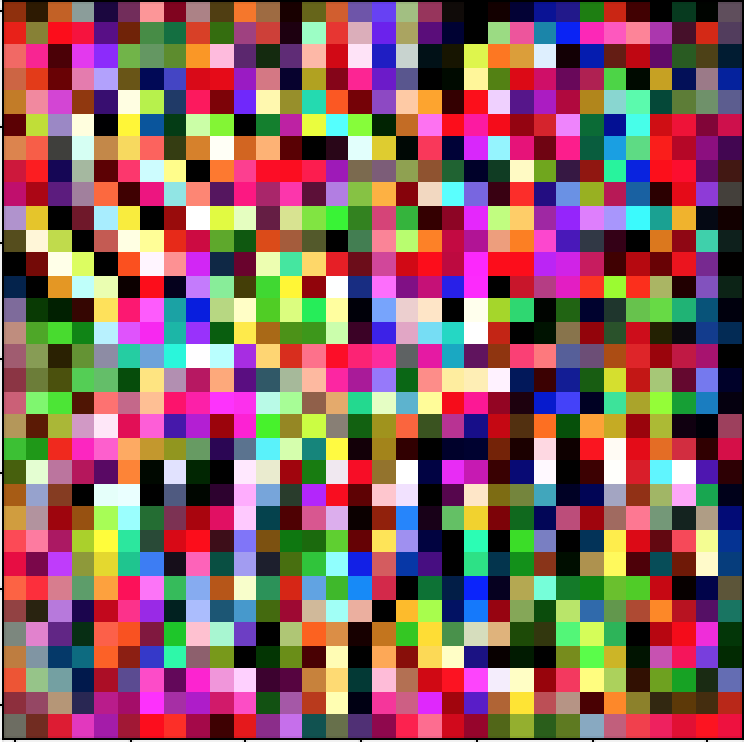}
 }  
\end{tabular}
 \\
  \begin{tabular}{@{\hskip 0.02in}c@{\hskip 0.02in} }
 \parbox[c]{10em}{\includegraphics[width=10em]{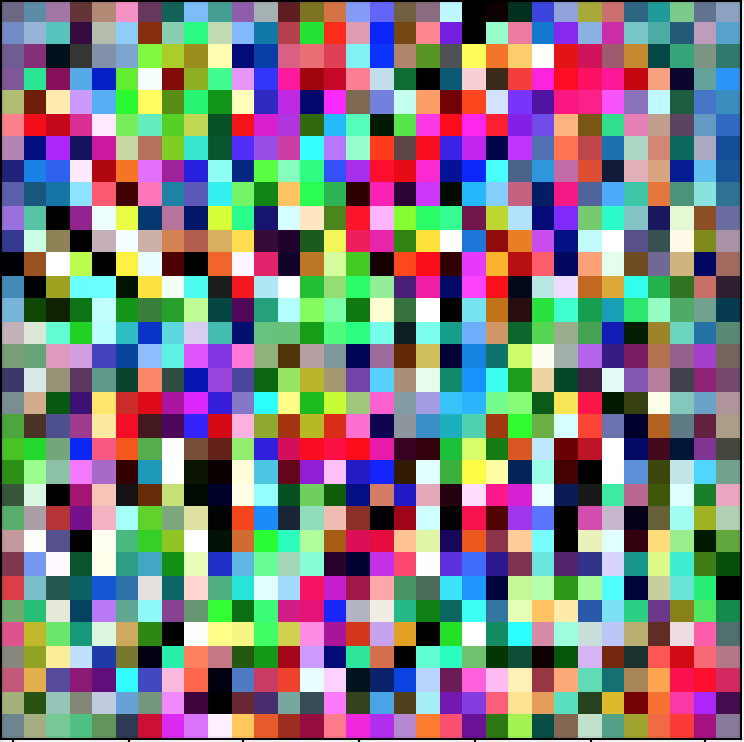}
 }  
\end{tabular}
 \\
 
\end{tabular}
&
 \begin{tabular}{@{}c@{}}  
\vspace*{-0.1in}\\
 \begin{tabular}{@{\hskip 0.02in}c@{\hskip 0.02in} }
 \parbox[c]{10em}{\includegraphics[width=10em]{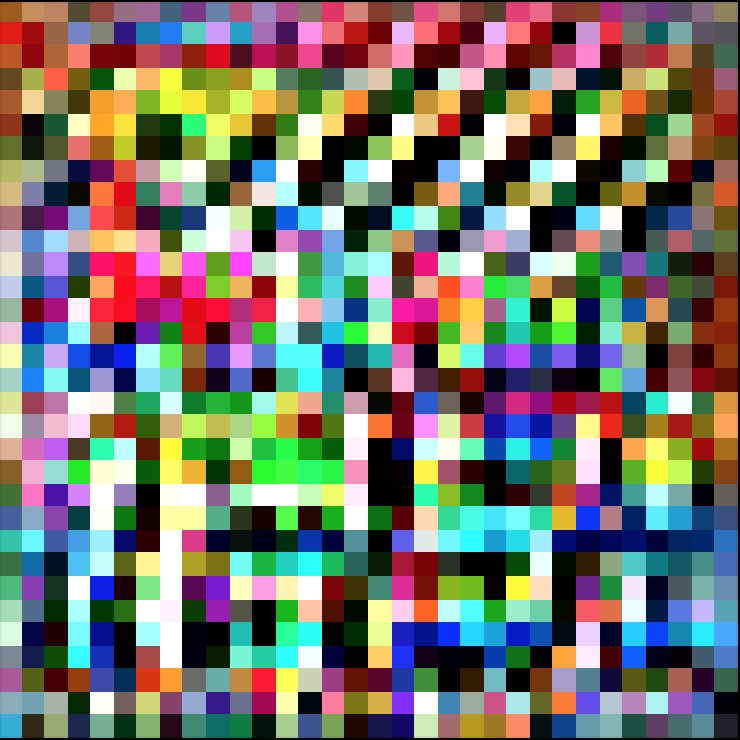}
 }  
\end{tabular} 
 \\
 \begin{tabular}{@{\hskip 0.02in}c@{\hskip 0.02in} }
 \parbox[c]{10em}{\includegraphics[width=10em]{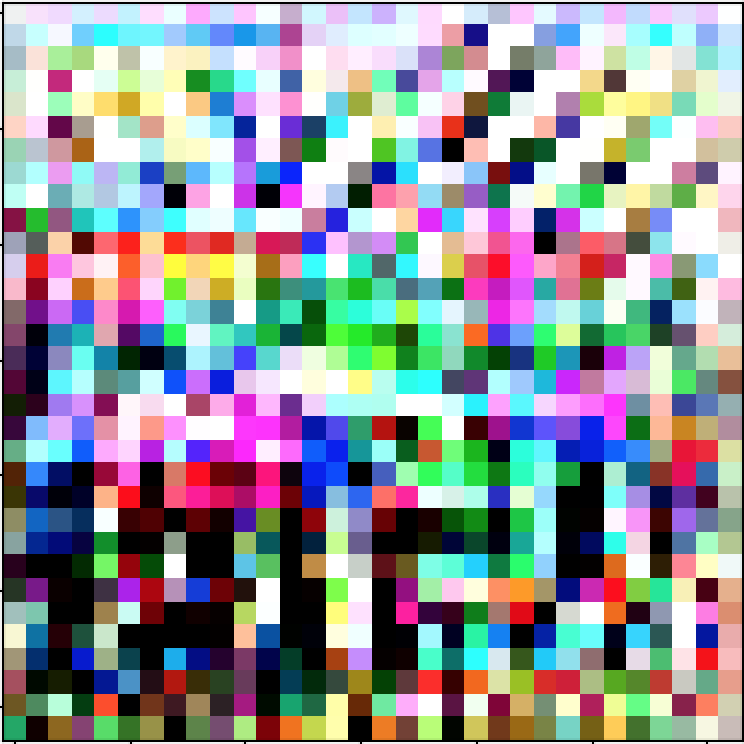}
 }  
\end{tabular}
 \\
 \begin{tabular}{@{\hskip 0.02in}c@{\hskip 0.02in} }
 \parbox[c]{10em}{\includegraphics[width=10em]{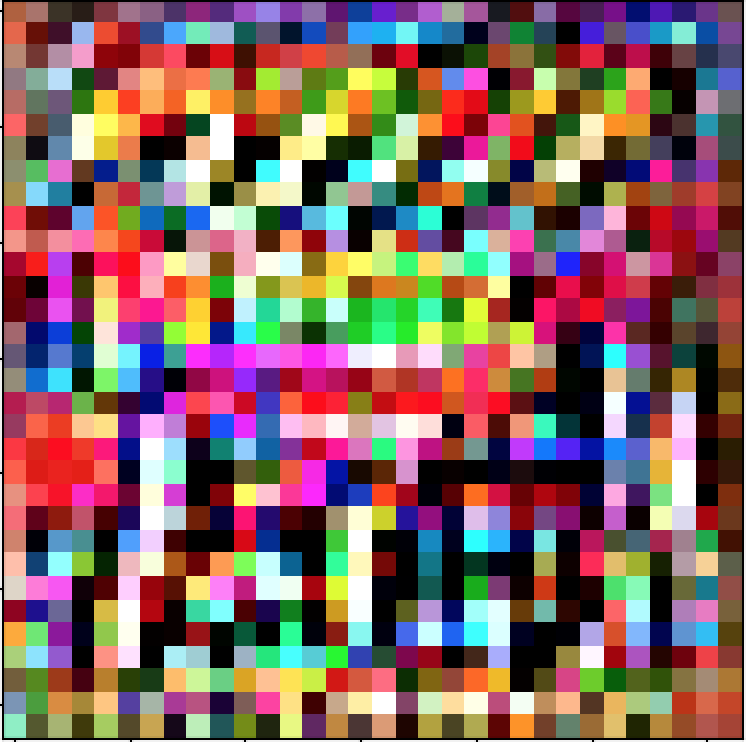}
 }  
\end{tabular}
 \\
  \begin{tabular}{@{\hskip 0.02in}c@{\hskip 0.02in} }
 \parbox[c]{10em}{\includegraphics[width=10em]{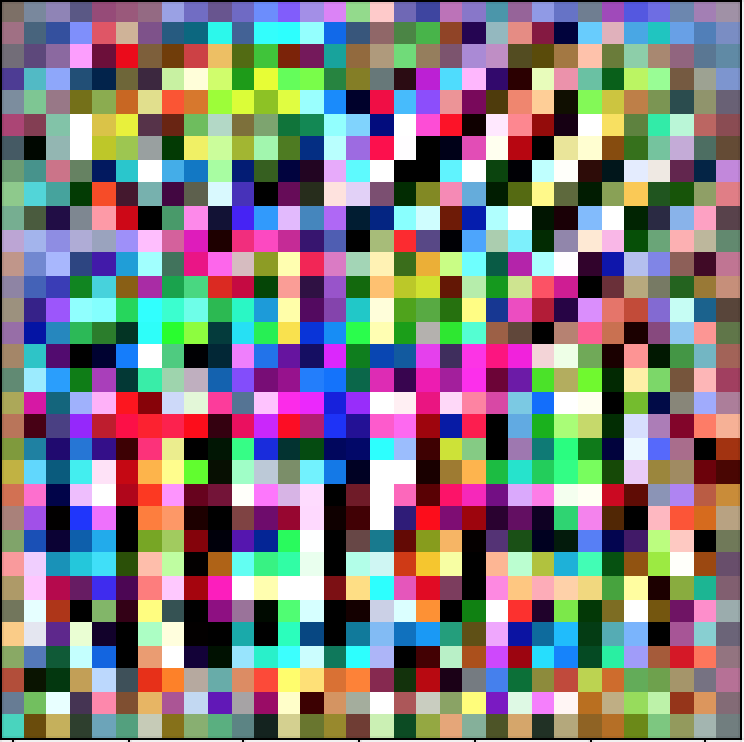}
 }  
\end{tabular}
 \\
 
\end{tabular} 
&
 \begin{tabular}{@{}c@{}}  
\vspace*{-0.1in}\\

 \begin{tabular}{@{\hskip 0.02in}c@{\hskip 0.02in} }
 \parbox[c]{10em}{\includegraphics[width=10em]{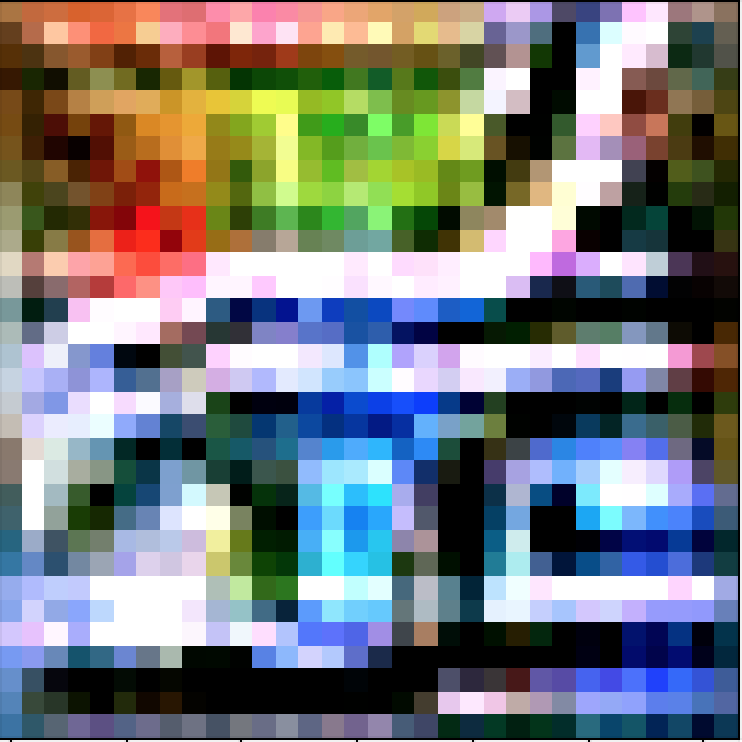}

 }  
\end{tabular} 
 \\
 \begin{tabular}{@{\hskip 0.02in}c@{\hskip 0.02in} }
 \parbox[c]{10em}{\includegraphics[width=10em]{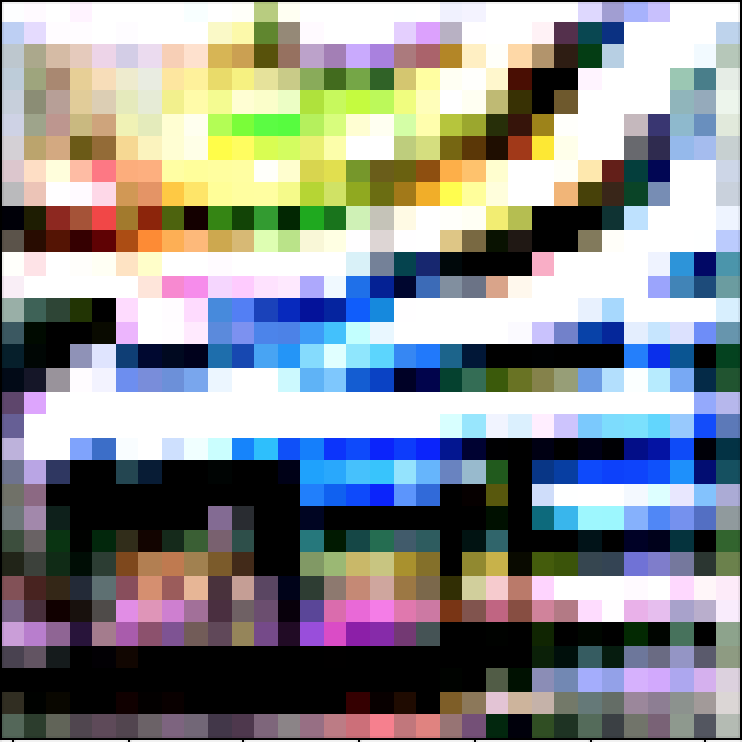}
 }  
\end{tabular}
 \\

 \begin{tabular}{@{\hskip 0.02in}c@{\hskip 0.02in} }
 \parbox[c]{10em}{\includegraphics[width=10em]{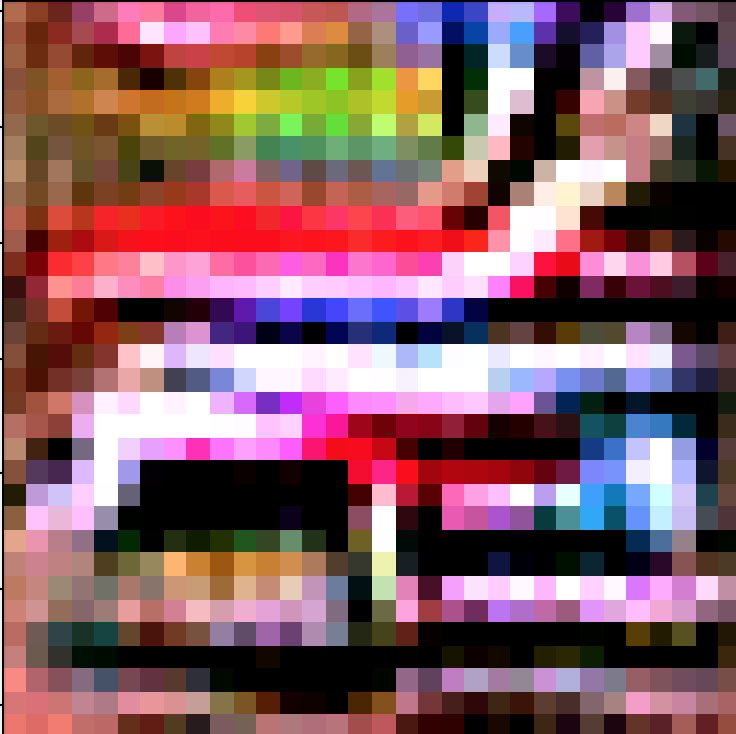}

 }  
\end{tabular}
 \\
 \begin{tabular}{@{\hskip 0.02in}c@{\hskip 0.02in} }
 \parbox[c]{10em}{\includegraphics[width=10em]{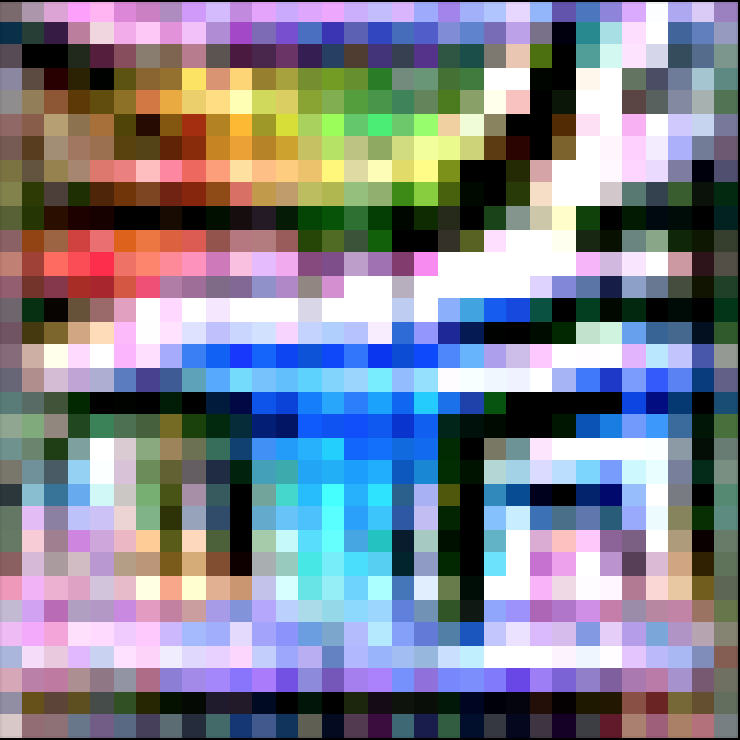}

 }  
\end{tabular}
 \\

\end{tabular} 
&
 \begin{tabular}{@{}c@{}}  
\vspace*{-0.1in}\\

 \begin{tabular}{@{\hskip 0.02in}c@{\hskip 0.02in} }
 \parbox[c]{10em}{\includegraphics[width=10em]{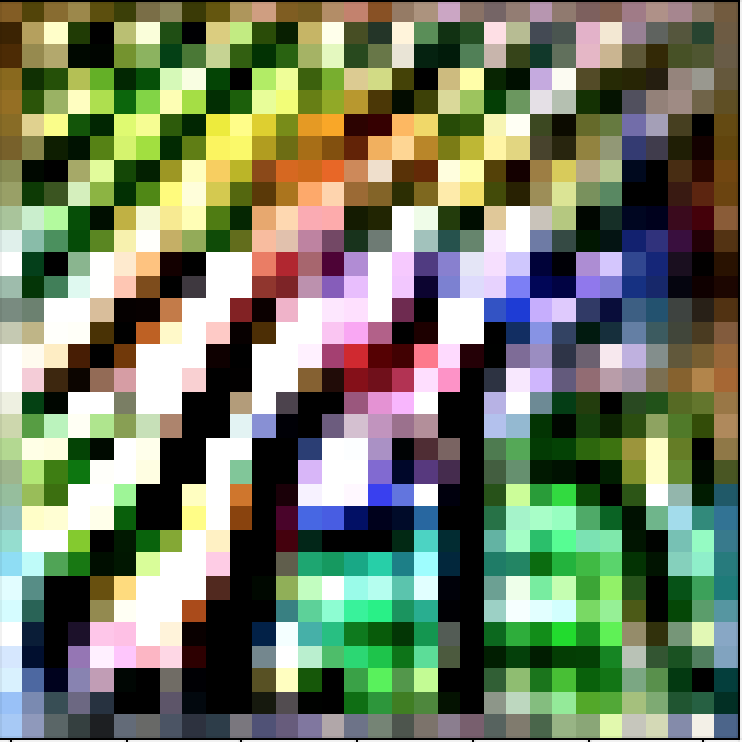}

 }  
\end{tabular} 
 \\

 \begin{tabular}{@{\hskip 0.02in}c@{\hskip 0.02in} }
 \parbox[c]{10em}{\includegraphics[width=10em]{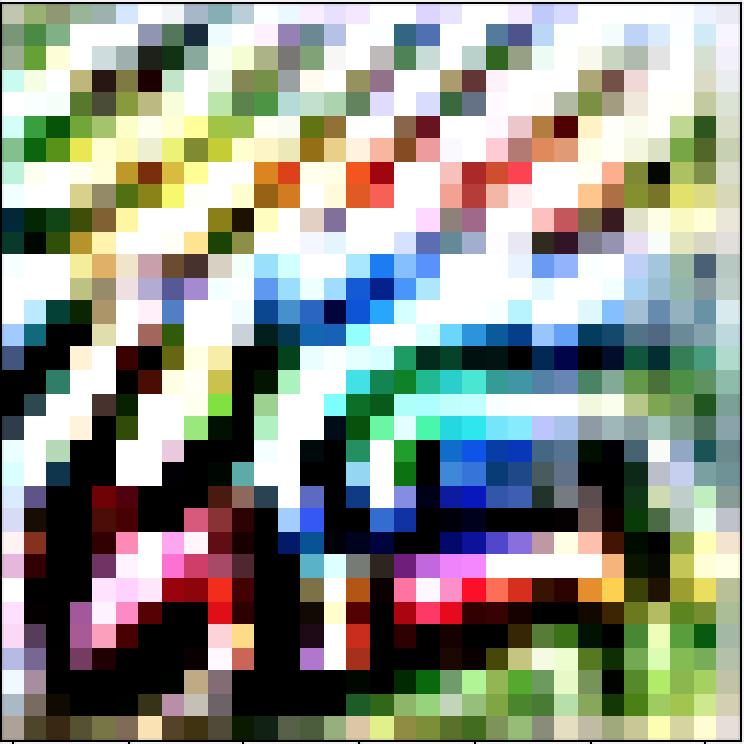}

 }  
\end{tabular}

 \\
 \begin{tabular}{@{\hskip 0.02in}c@{\hskip 0.02in} }
 \parbox[c]{10em}{\includegraphics[width=10em]{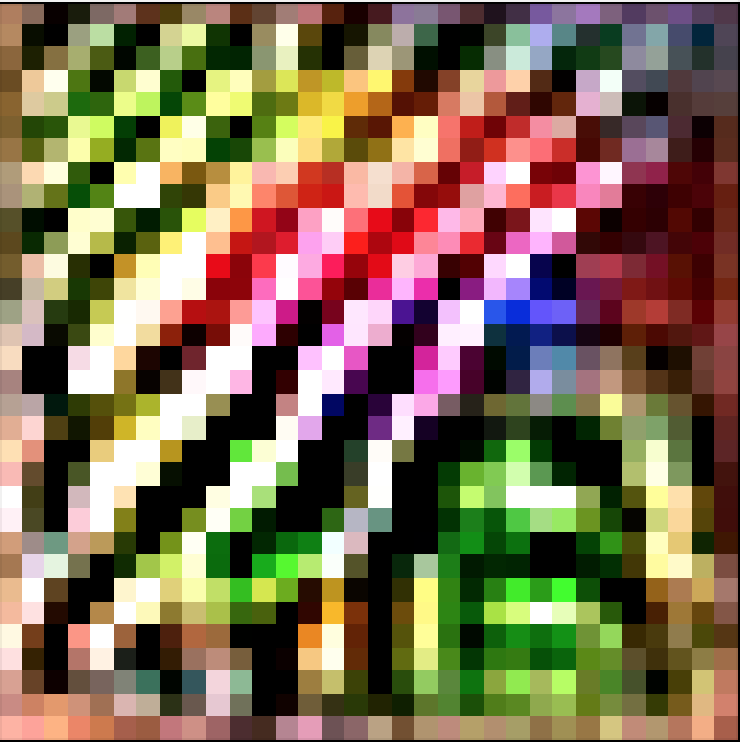}
 }  
\end{tabular}
 \\
 \begin{tabular}{@{\hskip 0.02in}c@{\hskip 0.02in} }
 \parbox[c]{10em}{\includegraphics[width=10em]{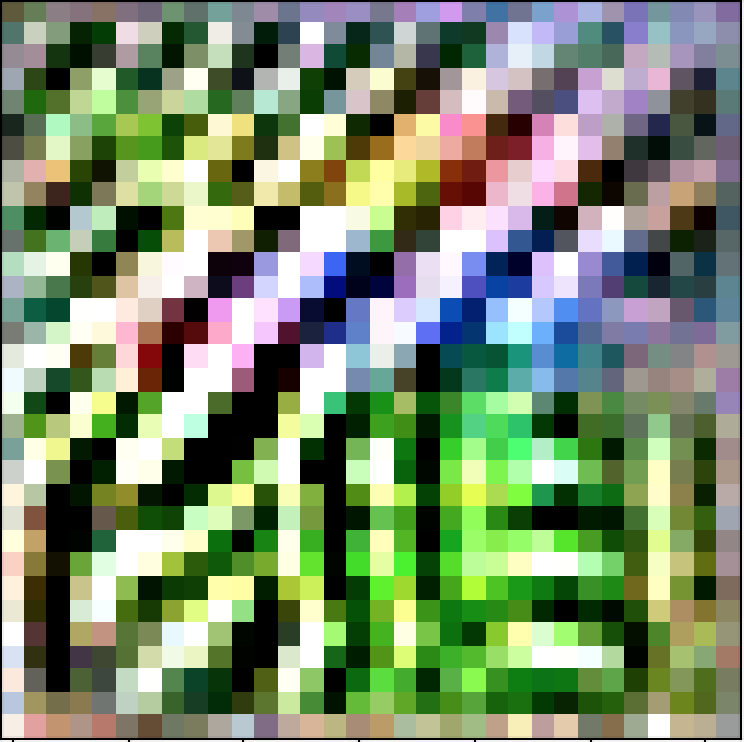}
 }  
\end{tabular}
 \\

\end{tabular} 
&
 \begin{tabular}{@{}c@{}}  
\vspace*{-0.1in}\\

 \begin{tabular}{@{\hskip 0.02in}c@{\hskip 0.02in} }
 \parbox[c]{10em}{\includegraphics[width=10em]{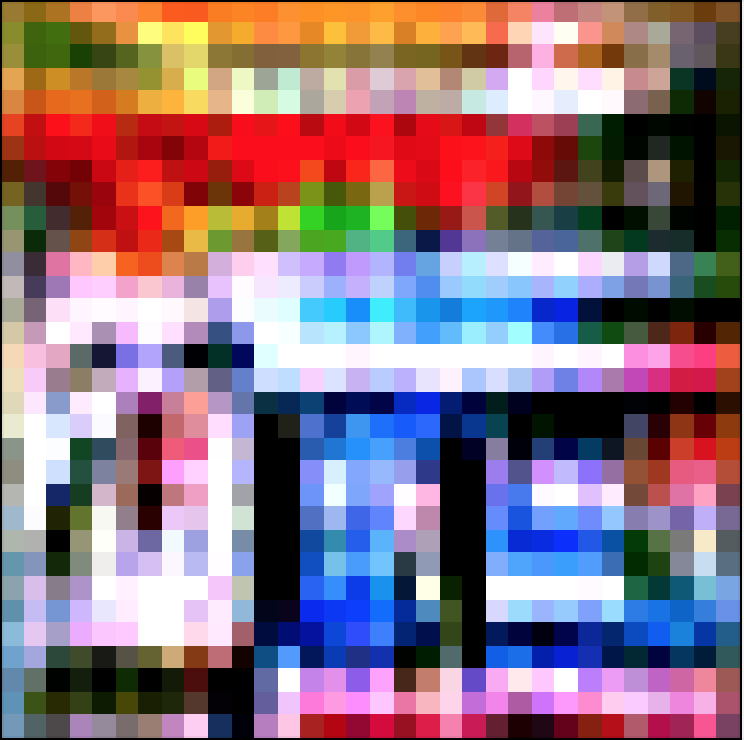}

 }  
\end{tabular} 
 \\
 \begin{tabular}{@{\hskip 0.02in}c@{\hskip 0.02in} }
 \parbox[c]{10em}{\includegraphics[width=10em]{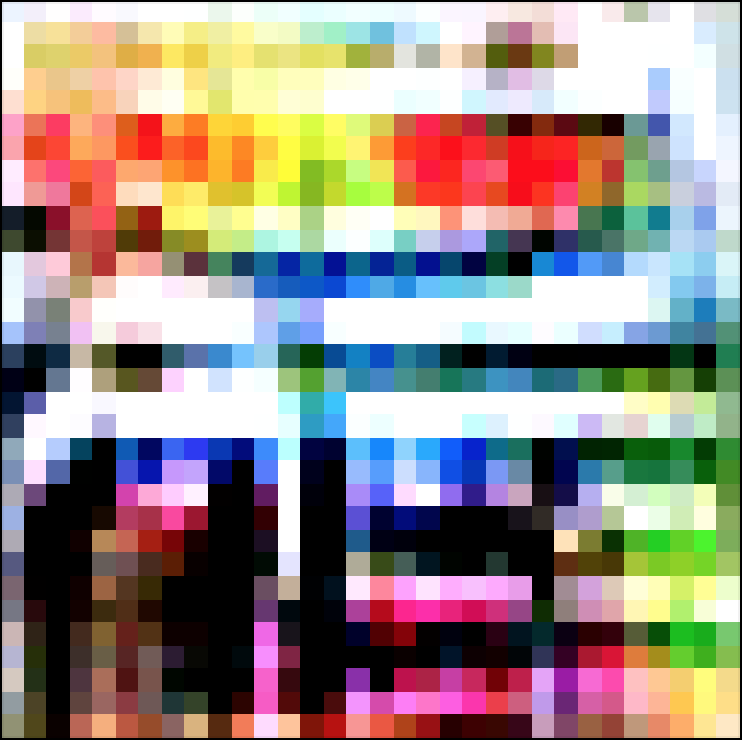}
 }  
\end{tabular}
 \\

 \begin{tabular}{@{\hskip 0.02in}c@{\hskip 0.02in} }
 \parbox[c]{10em}{\includegraphics[width=10em]{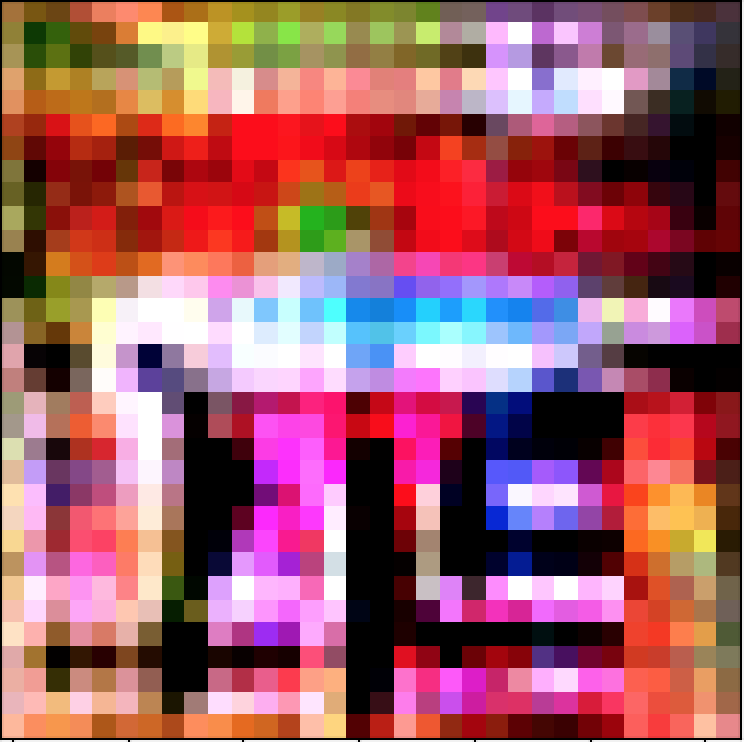}
 }  
\end{tabular}
\\
  \begin{tabular}{@{\hskip 0.02in}c@{\hskip 0.02in} }
 \parbox[c]{10em}{\includegraphics[width=10em]{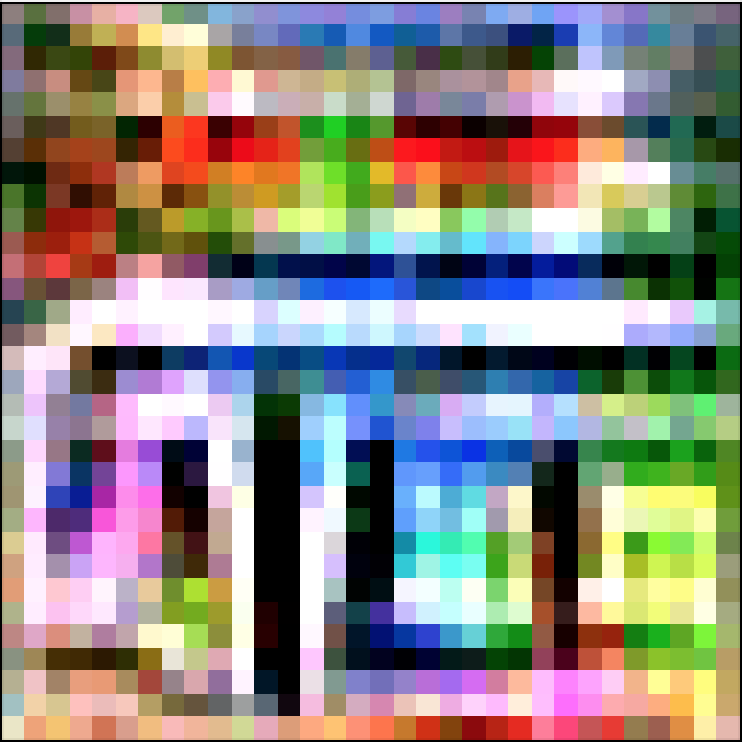}
 }  
\end{tabular}
 \\

\end{tabular} 

\end{tabular}
  \end{adjustbox}
\caption{\footnotesize{
Feature visualization at neuron $55$ under CIFAR-10 \texttt{Small} model trained by different methods.}}
\label{fig: robust_feature_cifar_small_56}
\end{figure}

 \begin{figure}[htb]
  \centering
  \begin{adjustbox}{max width=0.65\textwidth }
  \begin{tabular}{@{\hskip 0.00in}c @{\hskip 0.01in} | @{\hskip 0.01in}c   @{\hskip 0.00in}   @{\hskip 0.00in} c @{\hskip 0.00in}   @{\hskip 0.00in} c @{\hskip 0.00in}  @{\hskip 0.0in} c @{\hskip 0.00in}    @{\hskip 0.0in} c
  }
\colorbox{lightgray}{ \textbf{\Large{Seed Images}}} 
&
\textcolor{blue}{ \textbf{\Large{Normal}}}
&

\textcolor{blue}{\Large{ \textbf{Adv}} }
& 

\textcolor{blue}{\Large{  \textbf{Int}}}
& 
\textcolor{blue}{ \Large{  \textbf{Int2}} }
\\
 \begin{tabular}{@{}c@{}}  
\vspace*{-0.1in}\\

 \begin{tabular}{@{\hskip 0.02in}c@{\hskip 0.02in} }
 \parbox[c]{10em}{\includegraphics[width=10em]{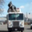}
 }  
\end{tabular} 
 \\
 \begin{tabular}{@{\hskip 0.02in}c@{\hskip 0.02in} }
 \parbox[c]{10em}{\includegraphics[width=10em]{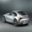}
 }  
\end{tabular}
 \\
 \begin{tabular}{@{\hskip 0.02in}c@{\hskip 0.02in} }
 \parbox[c]{10em}{\includegraphics[width=10em]{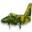}
 }  
\end{tabular}
 \\
 \begin{tabular}{@{\hskip 0.02in}c@{\hskip 0.02in} }
 \parbox[c]{10em}{\includegraphics[width=10em]{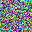}
 }  
\end{tabular}

\end{tabular} 
&
 \begin{tabular}{@{}c@{}}  
\vspace*{-0.1in}\\
 \begin{tabular}{@{\hskip 0.02in}c@{\hskip 0.02in} }
 \parbox[c]{10em}{\includegraphics[width=10em]{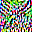}
 }  
\end{tabular} 
 \\
 \begin{tabular}{@{\hskip 0.02in}c@{\hskip 0.02in} }
 \parbox[c]{10em}{\includegraphics[width=10em]{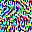}
 }  
\end{tabular}
 \\
 \begin{tabular}{@{\hskip 0.02in}c@{\hskip 0.02in} }
 \parbox[c]{10em}{\includegraphics[width=10em]{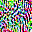}
 }  
\end{tabular}
 \\
  \begin{tabular}{@{\hskip 0.02in}c@{\hskip 0.02in} }
 \parbox[c]{10em}{\includegraphics[width=10em]{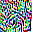}
 }  
\end{tabular}
 \\
 
\end{tabular}

&
 \begin{tabular}{@{}c@{}}  
\vspace*{-0.1in}\\

 \begin{tabular}{@{\hskip 0.02in}c@{\hskip 0.02in} }
 \parbox[c]{10em}{\includegraphics[width=10em]{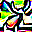}

 }  
\end{tabular} 
 \\
 \begin{tabular}{@{\hskip 0.02in}c@{\hskip 0.02in} }
 \parbox[c]{10em}{\includegraphics[width=10em]{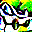}
 }  
\end{tabular}
 \\

 \begin{tabular}{@{\hskip 0.02in}c@{\hskip 0.02in} }
 \parbox[c]{10em}{\includegraphics[width=10em]{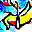}

 }  
\end{tabular}
 \\
 \begin{tabular}{@{\hskip 0.02in}c@{\hskip 0.02in} }
 \parbox[c]{10em}{\includegraphics[width=10em]{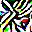}

 }  
\end{tabular}
 \\

\end{tabular} 
&
 \begin{tabular}{@{}c@{}}  
\vspace*{-0.1in}\\

 \begin{tabular}{@{\hskip 0.02in}c@{\hskip 0.02in} }
 \parbox[c]{10em}{\includegraphics[width=10em]{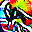}

 }  
\end{tabular} 
 \\

 \begin{tabular}{@{\hskip 0.02in}c@{\hskip 0.02in} }
 \parbox[c]{10em}{\includegraphics[width=10em]{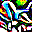}

 }  
\end{tabular}

 \\
 \begin{tabular}{@{\hskip 0.02in}c@{\hskip 0.02in} }
 \parbox[c]{10em}{\includegraphics[width=10em]{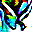}
 }  
\end{tabular}
 \\
 \begin{tabular}{@{\hskip 0.02in}c@{\hskip 0.02in} }
 \parbox[c]{10em}{\includegraphics[width=10em]{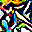}
 }  
\end{tabular}
 \\

\end{tabular} 
&
 \begin{tabular}{@{}c@{}}  
\vspace*{-0.1in}\\

 \begin{tabular}{@{\hskip 0.02in}c@{\hskip 0.02in} }
 \parbox[c]{10em}{\includegraphics[width=10em]{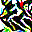}

 }  
\end{tabular} 
 \\
 \begin{tabular}{@{\hskip 0.02in}c@{\hskip 0.02in} }
 \parbox[c]{10em}{\includegraphics[width=10em]{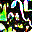}
 }  
\end{tabular}
 \\

 \begin{tabular}{@{\hskip 0.02in}c@{\hskip 0.02in} }
 \parbox[c]{10em}{\includegraphics[width=10em]{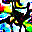}
 }  
\end{tabular}
\\
  \begin{tabular}{@{\hskip 0.02in}c@{\hskip 0.02in} }
 \parbox[c]{10em}{\includegraphics[width=10em]{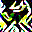}
 }  
\end{tabular}
 \\

\end{tabular} 

\end{tabular}
  \end{adjustbox}
\caption{\footnotesize{
Feature visualization at neuron $42$ under CIFAR-10 \texttt{WResnet} model trained by different methods.
}}
\label{fig: robust_feature_cifar_resnet_56}
\end{figure}

 \begin{figure}[htb]
  \centering
  \begin{adjustbox}{max width=0.65\textwidth }
  \begin{tabular}{@{\hskip 0.00in}c @{\hskip 0.01in} | @{\hskip 0.01in}c   @{\hskip 0.00in}   @{\hskip 0.00in} c @{\hskip 0.00in}   @{\hskip 0.00in} c @{\hskip 0.00in}  @{\hskip 0.0in} c @{\hskip 0.00in}    @{\hskip 0.0in} c 
  }
\colorbox{lightgray}{ \textbf{\Large{Seed Images}}} 
&
\textcolor{blue}{ \textbf{\Large{Normal}}}
&

\textcolor{blue}{\Large{ \textbf{Adv}} }
& 

\textcolor{blue}{\Large{  \textbf{Int}}}
& 
\textcolor{blue}{ \Large{  \textbf{Int2}} }
\\
 \begin{tabular}{@{}c@{}}  
\vspace*{-0.1in}\\

 \begin{tabular}{@{\hskip 0.02in}c@{\hskip 0.02in} }
 \parbox[c]{10em}{\includegraphics[width=10em]{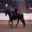}
 }  
\end{tabular} 
 \\
 \begin{tabular}{@{\hskip 0.02in}c@{\hskip 0.02in} }
 \parbox[c]{10em}{\includegraphics[width=10em]{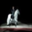}
 }  
\end{tabular}
 \\
 \begin{tabular}{@{\hskip 0.02in}c@{\hskip 0.02in} }
 \parbox[c]{10em}{\includegraphics[width=10em]{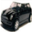}
 }  
\end{tabular}
 \\
 \begin{tabular}{@{\hskip 0.02in}c@{\hskip 0.02in} }
 \parbox[c]{10em}{\includegraphics[width=10em]{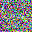}
 }  
\end{tabular}

\end{tabular} 
&
 \begin{tabular}{@{}c@{}}  
\vspace*{-0.1in}\\
 \begin{tabular}{@{\hskip 0.02in}c@{\hskip 0.02in} }
 \parbox[c]{10em}{\includegraphics[width=10em]{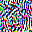}
 }  
\end{tabular} 
 \\
 \begin{tabular}{@{\hskip 0.02in}c@{\hskip 0.02in} }
 \parbox[c]{10em}{\includegraphics[width=10em]{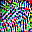}
 }  
\end{tabular}
 \\
 \begin{tabular}{@{\hskip 0.02in}c@{\hskip 0.02in} }
 \parbox[c]{10em}{\includegraphics[width=10em]{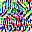}
 }  
\end{tabular}
 \\
  \begin{tabular}{@{\hskip 0.02in}c@{\hskip 0.02in} }
 \parbox[c]{10em}{\includegraphics[width=10em]{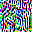}
 }  
\end{tabular}
 \\
 
\end{tabular}

&
 \begin{tabular}{@{}c@{}}  
\vspace*{-0.1in}\\

 \begin{tabular}{@{\hskip 0.02in}c@{\hskip 0.02in} }
 \parbox[c]{10em}{\includegraphics[width=10em]{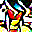}

 }  
\end{tabular} 
 \\
 \begin{tabular}{@{\hskip 0.02in}c@{\hskip 0.02in} }
 \parbox[c]{10em}{\includegraphics[width=10em]{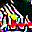}
 }  
\end{tabular}
 \\

 \begin{tabular}{@{\hskip 0.02in}c@{\hskip 0.02in} }
 \parbox[c]{10em}{\includegraphics[width=10em]{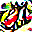}

 }  
\end{tabular}
 \\
 \begin{tabular}{@{\hskip 0.02in}c@{\hskip 0.02in} }
 \parbox[c]{10em}{\includegraphics[width=10em]{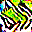}

 }  
\end{tabular}
 \\

\end{tabular} 
&
 \begin{tabular}{@{}c@{}}  
\vspace*{-0.1in}\\

 \begin{tabular}{@{\hskip 0.02in}c@{\hskip 0.02in} }
 \parbox[c]{10em}{\includegraphics[width=10em]{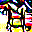}

 }  
\end{tabular} 
 \\

 \begin{tabular}{@{\hskip 0.02in}c@{\hskip 0.02in} }
 \parbox[c]{10em}{\includegraphics[width=10em]{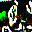}

 }  
\end{tabular}

 \\
 \begin{tabular}{@{\hskip 0.02in}c@{\hskip 0.02in} }
 \parbox[c]{10em}{\includegraphics[width=10em]{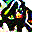}
 }  
\end{tabular}
 \\
 \begin{tabular}{@{\hskip 0.02in}c@{\hskip 0.02in} }
 \parbox[c]{10em}{\includegraphics[width=10em]{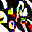}
 }  
\end{tabular}
 \\

\end{tabular} 
&
 \begin{tabular}{@{}c@{}}  
\vspace*{-0.1in}\\

 \begin{tabular}{@{\hskip 0.02in}c@{\hskip 0.02in} }
 \parbox[c]{10em}{\includegraphics[width=10em]{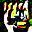}

 }  
\end{tabular} 
 \\
 \begin{tabular}{@{\hskip 0.02in}c@{\hskip 0.02in} }
 \parbox[c]{10em}{\includegraphics[width=10em]{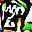}
 }  
\end{tabular}
 \\

 \begin{tabular}{@{\hskip 0.02in}c@{\hskip 0.02in} }
 \parbox[c]{10em}{\includegraphics[width=10em]{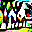}
 }  
\end{tabular}
\\
  \begin{tabular}{@{\hskip 0.02in}c@{\hskip 0.02in} }
 \parbox[c]{10em}{\includegraphics[width=10em]{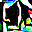}
 }  
\end{tabular}
 \\

\end{tabular} 

\end{tabular}
  \end{adjustbox}
\caption{\footnotesize{
Feature visualization at neuron $3$ under CIFAR-10  \texttt{WResnet} model trained by different methods. 
}}
\label{fig: robust_feature_cifar_wide_1}
\end{figure}

\end{document}